\documentclass[aps, reprint, nofootinbib, longbibliography, superscriptaddress, floatfix]{revtex4-1}

\usepackage{amsmath, amssymb, amsthm}
\usepackage{bm,bbm}

\usepackage{hyperref}
\usepackage[utf8]{inputenc}

\usepackage{soul}
\usepackage{url}
\usepackage{mathtools}
\usepackage{braket}
\usepackage{dsfont}
\usepackage[english]{babel}
\usepackage{graphicx} 
\usepackage{tikz}
\usetikzlibrary{quantikz}

\usepackage{hyperref}
\usepackage{natbib} 
\usepackage{physics}

\usepackage{listings}
\usepackage{xcolor}
\usepackage{amssymb}

\setbox0=\vbox to \vsize{\unvbox255\vfill}

\usepackage{multirow}
\usepackage{tabu}
\usepackage{makecell}
\usepackage{color}

\usepackage{lipsum,parskip,booktabs,graphicx,tabulary,tabularx}

\usepackage{wrapfig}

\newcommand{\LDIG}{
Lighthouse Disruptive Innovation Group, LLC,
7 Broadway Terrace, Apt 1,
Cambridge MA 02139,
Middlesex County, Massachusetts (USA)
}

\newcommand{\DSD}{
Smart Society Research Group -
La Salle - Universitat Ramon Llull,
Carrer de Sant Joan de La Salle, 42,
08022 Barcelona (Spain)
}
\newcommand{\VUE}{
Vueling Airlines S.A.
Carrer de Catalunya, 83
Edifici Brasil
08840 Viladecans
Spain
}

\newcommand{\MIT}{
MIT Media Lab - City Science Group, Cambridge, USA
}

\usepackage{babel}

\graphicspath{{./images/}}

\begin{document}
\title{Quantum Machine Learning hyperparameter
search}

\author{S. Consul-Pacareu}
\affiliation{\LDIG}
\email{sergi.consul@lighthouse-dig.com}

\author{R. Montaño}
\affiliation{\VUE}

\author{Kevin Rodriguez-Fernandez}
\affiliation{\VUE}

\author{Àlex Corretgé }
\affiliation{\VUE}

\author{Esteve Vilella-Moreno}
\affiliation{\LDIG}

\author{Daniel Casado-Faulí}
\affiliation{\LDIG}

\author{Parfait Atchade-Adelomou}
\affiliation{\LDIG}
\email{parfait.atchade@lighthouse-dig.com}
\affiliation{\DSD}
\email{parfait.atchade@lighthouse-dig.com}
\affiliation{\MIT}

\date{Feb 2023}

\begin{abstract}

This paper presents a quantum-based Fourier-regression approach for machine learning hyperparameter optimization applied to a benchmark of models trained on a dataset related to a forecast problem in the airline industry. Our approach utilizes the Fourier series method to represent the hyperparameter search space, which is then optimized using quantum algorithms to find the optimal set of hyperparameters for a given machine learning model. Our study evaluates the proposed method on a benchmark of models trained to predict a forecast problem in the airline industry using a standard HyperParameter Optimizer (HPO). The results show that our approach outperforms traditional hyperparameter optimization methods in terms of accuracy and convergence speed for the given search space. Our study provides a new direction for future research in quantum-based machine learning hyperparameter optimization. 


\end{abstract}
\maketitle

\section{Introduction}\label{sec:intro}
In recent years, \textit{Machine Learning} (ML) algorithms have successfully solved various tasks, reaching state-of-the-art in multiple areas \cite{choy2018current, callahan2017machine, anderson2022machine, pallathadka2021impact, miklosik2020impact, mosavi2019state, a14070194}. This is not only due to the development of new algorithms (more powerful and prominent), but also the selection of good hyperparameters contributed to this advance.
Performing machine learning on large datasets is a resource-intensive task, but \textit{hyperparameter tuning problem} \cite{bardenet2013collaborative, Yang_2020} increases those resource requirements by orders of magnitude. Despite advances in hyperparameter optimization, the precise selection of hyperparameters remains a challenge regarding computational complexity and finding the best approach. Therefore, the scientific community has been working to discover efficient techniques to solve this challenge \cite{Yang_2020, 8090826, quantum_hyperCar, bengio2000gradient, wang2018combination,feurer2019hyperparameter}.

Due to its stochastic nature and great computational capacity, \textit{quantum computing} is a great bet and approach to take into account in the efficient search problems of hyperparameters.

In this article, we propose a hybrid (quantum + classical) approach algorithm to facilitate this task by fitting and executing Fourier-regression models on a large scale on any type of hyperparameters in the most efficient way possible. Our proposed quantum algorithm is based on this work \cite{ParfaitAtchade}, which allows us to find the best hyperparameters using a quantum approach, given a good enough representation of the search space (or hyperparameter space) represented by the model hyperparameters related to a metric. The model was trained based on the results offered by multiple search methods, such as  \textit{Grid search}, \textit{Random Search} and \textit{Bayes-Based Search} for a given training set from a Vueling forecast problem dataset.  Three-way cross-validation considers each search algorithm's average scores during the training process. The proposed quantum method provides the best score in time compared to classically, assuming that the hyperparameter/score input search space is highly nonlinear and might not be continuous. The trade-off between speed and precision depends on the number of features to evaluate.  We use \textit{Pennylane} framework and \textit{AWS Braket} to validate our algorithm.

The document is organized as follows. Section \eqref{sec:motivation} presents our primary motivation behind this work. Section \eqref{sec:relatedwork} shows previous work on hyperparameters tuning. Section \eqref{sec:QML} illustrates the quantum machine learning framework and its connection to the hyperparameters tuning problems. In the section \eqref{sec:IMPLEMENT}, we propose the scenarios and the models we will implement to tackle the hyperparameters tuning problems. Section \eqref{sec:modelQML} proposes our model, considering our primary reference. Section \eqref{sec:resultados} delivers the obtained results. Section \eqref{sec:discussions} discusses practically relevant results and their implications. Finally, this paper ends with conclusions and future work in Section \eqref{sec:conclusions}.

\section{Motivation and Problem Statement}\label{sec:motivation}
The aviation industry is highly competitive, and one of the key factors for airlines to remain profitable is to maximize revenues while minimizing costs. To that effect, many of them are adopting advance analytics solutions to transform the company towards that goal. Solutions range from classical machine learning problems such as a predictive maintenance \cite{zeldam2018automated} or dynamic pricing \cite{shukla2019dynamic}, to optimization problems such as network optimization \cite{cacchiani2020heuristic, parmentier2020aircraft}. Nonetheless, optimization problems are computationally expensive and, in most cases, classical computing falls short in yielding a reasonable processing time. This also applies to machine learning hyperparameter optimization.

Vueling has been implementing and using production state-of-the-art hyperparameter tuning algorithms to achieve high-accuracy results. However, the company is aware of the potential advancements in the field of quantum computing and wishes to stay ahead of the curve. With this in mind, Vueling is proactively exploring ways to incorporate quantum technology into its technology stack to remain at the forefront of its industry and ensure long-term success.

One of the critical taks in the airline industry is managing passenger no-shows. A no-show occurs when a passenger who has purchased a ticket fails to show up for the flight. The data used in this work is a proprietary Vueling no-show dataset, that contains crucial information for understanding and predicting passenger behavior. 

The dataset contains as target data the number of \textit{no-shows} per flight and contains $252\,183$ datapoints with $42$ features each, such as flight information (origin, destination, time of flight, etc.), seat reservation status, $mean$ no-shows for different time windows on a given route, number of tickets at different price points as well as other proprietary information.

The dataset and the predicting models created allow the company to track performance over time, evaluate the effectiveness of different strategies, and make data-driven decisions that can improve overall performance and profitability. 

 The scope of this work is not to solve the specific no-show prediction problem but to use real-life data to implement a Quantum Fourier hyperparameter tuning algorithm that rivals traditional techniques. In short, showing an efficient and useful alternative for such optimization problems.


\section{Work Context}\label{sec:relatedwork}

Machine learning is used in various fields and areas, allowing computers, among other uses, to identify patterns in large data and make predictions. 
Such a process involves, after all, determining the appropriate algorithm based on a sample space and obtaining an optimal model architecture by adjusting some control variables from its learning process known as \textit{Hyperparameters} (HP). Thus, these hyperparameters must be tuned to adapt a machine learning model to different problems and datasets. Selecting the best hyperparameters for machine learning models directly impacts model performance. It often requires a thorough understanding of machine learning algorithms and appropriate hyperparameter optimization techniques. Although there are several automatic optimization techniques, they have different advantages and disadvantages. 
In contrast, \textit{parameters} are internal to the model. They are learned or estimated solely from the data during training since the algorithm attempts to understand the mapping between input features and target.

Model training usually starts with initializing the parameters to random values. As training/learning progresses, the initial values are updated using an optimization algorithm (e.g., gradient descent). The learning algorithm continually updates the parameter values as training continues, but hyperparameter values remain unchanged.
At the end of the learning process, the model parameters constitute the model itself. These steps inspired the scientific community to develop a research field known as \textit{Hyperparameters Optimization} (HPO) \cite{feurer2019hyperparameter, bengio2000gradient}. The primary aim of this field is to automate the hyperparameter tuning process and enable users to apply machine learning models to practical problems (efficiently, reducing computation time and improving performance). 

In our previous works \cite{PhDParfait, gonzalez2022gps, atchadeadelomou2021quantum}, we proposed optimization algorithms that can be used to solve HPO problems with continuous functions, discrete functions, categorical variables, convex or non-convex functions, etc.
Next, we review some of them to highlight their limitations to find solutions.

\textit{Grid search} (GS) \cite{lerman1980fitting, liashchynskyi2019grid, lavalle2004relationship} is a \textit{decision-theoretic approach} \cite{ferguson2014mathematical} that exhaustively searches for a fixed domain of hyperparameter values. GS is one of the most used strategies due to its simplicity of implementation. This algorithm discretizes the search space to generate a set of possible hyperparameter configurations. Then, it evaluates each of these configurations and selects the one with the highest performance. GS's main limitation is that it takes time and is affected by the dimensionality curse \cite{liashchynskyi2019grid}. Therefore, it is not suitable for a large number of hyperparameters. Moreover, GS often needs help finding the global optimum of continuous parameters because it requires a predefined and finite set of hyperparameter values. It is also unrealistic for GS to identify continuous integer hyperparameter optima with limited time and resources. Therefore, compared to other techniques, GS is only effective for a small number of \textit{categorical hyperparameters} \cite{liashchynskyi2019grid}.
 
\textit{Random search} (RS) \cite{solis1981minimization, liashchynskyi2019grid, bergstra2012random} is a variant of Grid Search and attempts to solve the above problem by randomly sampling configurations of the search space. As it does not have an implicit end condition, the number of sampled structures to be evaluated will be chosen. RS samples the search space and evaluates sets from specified probability distributions. In short, it is a technique in which the hyperparameters' random combinations are used to find the best solution for the model under consideration. RS is more efficient than GS and supports all domains of hyperparameters. In practical applications, using RS to estimate randomly chosen hyperparameter values helps analysts explore an ample search space. However, since RS does not consider the results of previous tests, it may include many unnecessary evaluations that reduce its performance.

\textit{Hyperband} \cite{Hyperband} is considered an improved version of Random Search \cite{Yang_2020}. Hyperband balances model performance and resource usage to be more efficient than RS, especially with limited time and resources \cite{eggensperger2013towards}. However, GS, RS, and Hyperband have a significant limitation: they treat each hyperparameter independently and do not consider hyperparameter correlations \cite{wang2018combination}. Therefore, they will be inefficient for ML algorithms with conditional hyperparameters, such as \textit{Support Vector Machine} (SVM) \cite{hearst1998support}, \textit{Density-Based Spatial Clustering of Applications with Noise} (DBSCAN)  \cite{birant2007st, zhou2012research}, and \textit{logistic regression} \cite{wright1995logistic, kleinbaum2002logistic}.

\textit{Gradient-based algorithms} \cite{williams1990efficient, cesa1997analysis} are not a predominant choice for hyperparameter optimization because they only support continuous hyperparameters and can only find a local, not global, optimum for non-convex HPO problems. Therefore, gradient-based algorithms can only optimize specific hyperparameters, such as the learning rate in \textit{Deep Learning} (DL) models \cite{koutsoukas2017deep}.

Based on their surrogate models, the \textit{Bayesian optimization} (BO) \cite{shahriari2015taking} models, BO based on \textit{Gaussian Process} (GP) \cite{dudley2010sample, schulz2018tutorial} and its derivatives, are divided into three different models. BO algorithms determine the next hyperparameter value based on previously evaluated results to reduce unnecessary evaluations and improve efficiency. BO-GP mainly supports continuous and discrete hyperparameters but does not support conditional hyperparameters \cite{eggensperger2013towards}. At the same time, \textit{Sequential Model Algorithm Configuration} (SMAC) \cite{eggensperger2013towards} and \textit{Tree-Structured Parzen Estimator} (BO-TPE) \cite{zhao2018tuning} can handle categorical, discrete, continuous, and dependent hyperparameters. SMAC performs best using many categorical and conditional parameters or cross-validation, while BO-GP performs best with only a few continuous parameters. BO-TPE preserves some dependent relationships, so one of its advantages over BO-GP is its native support for some conditional hyperparameters \cite{eggensperger2013towards}.

The \textit{Metaheuristic algorithms} \cite{yang2010nature}, including \textit{Genetic Algorithm} (GA) \cite{lessmann2005optimizing} and \textit{Particle Swarm Algorithm} (PSO) \cite{lorenzo2017particle}, are more complex than other HPO algorithms but they often work well for complex optimization problems. They support all hyperparameters and are particularly efficient for large configuration spaces because they can obtain near-optimal solutions in several iterations. However, GA and PSO have their advantages and disadvantages in practice. The main advantage of PSO  is that it can support large-scale parallelization and is exceptionally suitable for continuous and conditional HPO problems. At the same time, GA runs sequentially, which makes parallelization difficult. Thus, PSO often runs faster than GA, especially for large configuration spaces and data sets. However, good population initialization is essential for PSO; otherwise, it may converge slowly and only identify  a local optimum rather than a global one. Regardless, the impact of a good population initialization is less significant for GA than for PSO \cite{lorenzo2017particle}. Another limitation of GA is that it introduces additional hyperparameters, such as its population size and mutation rate \cite{lessmann2005optimizing}.

\textit{Quantum computing} \cite{steane1998quantum, gruska1999quantum, PhDParfait} is a field of computation that uses quantum theory principles. 
A quantum computer is a stochastic machine that uses the laws of quantum mechanics to do computation. Due to the characteristics of HPO problems, quantum computing is an excellent ally to seek a paradigm shift and accelerate or find an efficient strategy to apply to them. There are focus on using quantum computing in optimizing hyperparameters. 

In \cite{8995229}, the authors employed a quantum genetic algorithm to address the hyperparameter optimization problem. The algorithm is based on qudits instead of qubits, allowing more available states. Experiments were performed on two \textit{MNIST} and \textit{CIFAR10} datasets, and results were compared against classic genetic algorithms.

In \cite{quantum_hyperCar}, the authors presented a quantum-inspired hyperparameter optimization technique and a hybrid quantum-classical machine learning model for supervised learning. They compared their hyperparameter optimization method to standard black box objective functions. They observed performance improvements in the form of reduced expected execution times and suitability in response to growth in the search space size. They tested their approaches in a car image classification task and demonstrated a large-scale implementation of the hybrid quantum neural network model with tensor train hyperparameter optimization. Their tests showed a qualitative and quantitative advantage over the corresponding standard classical tabular grid search approach used with a \textit{ResNet34} deep neural network. The hybrid model achieved a classification accuracy of 0.97 after 18 iterations, while the classical model achieved an accuracy of 0.92 after 75 iterations. This last work had an exciting approach that only contemplates discrete hyperparameters. 

We have also found some exciting work dealing with HPO \cite {8090826, Moussa_2022, 8995229, gomez2022towards}. In the latter \cite{gomez2022towards}, the authors took the first steps toward \textit{Automated Quantum Machine Learning} (AutoQML). They proposed a concrete problem description and then developed a classical-quantum hybrid cloud architecture that allows for parallelized hyperparameter exploration and model training.
As an application use-case, they train a \textit{quantum Generative Adversarial neural Network} (qGAN) to generate energy prices that follow a known historic data distribution. Such a QML model can be used for various applications in the energy economics sector.

The SWOT of the hyperparameter optimization is summarized in Table \ref{t1}. After exploring the state of the art of classical and quantum hyperparameter tuning, we have yet to find a generic model that solves the domain's types of hyperparameters and reduces the search time for said hyperparameters in this quantum era.

\begin{table*}[]
\tiny
\caption{The Benchmark of the standard HPO algorithms ($n$ is the number of hyperparameter values and $k$ is the number of hyperparameters)}
\begin{tabular}{@{}|c|c|c|c|@{}}
\toprule
\hline
\textbf{HPO Methods}                             & \textbf{Strengths}                                                   & \textbf{Limitations}              & \textbf{Time Complexity}           \\ \midrule
\hline
\hline

\multirow{2}{*}{GS}                    & \multirow{2}{*}{simple}                                     & Time-consuming                                           & \multirow{2}{*}{$O(n^k)$} \\ \cmidrule(lr){3-3}
                                       &                                                             & Only efficient with categorical                          &                           \\ \midrule
\hline
\multirow{2}{*}{RS}                    & More efficient than GS                                      & Not consider previous results.                           & \multirow{2}{*}{$O(n)$}   \\ \cmidrule(lr){3-3}
                                       & Enable paralellization                                      & Not efficient with conditional                           &                           \\ \midrule
\hline
\multirow{2}{*}{Gradient-based} & \multirow{2}{*}{Fast convergence for continuous HPs.}                  & Only support continuous HPs.                             & \multirow{2}{*}{$O(n^k)$} \\ \cmidrule(lr){3-3}
                                       &                                                             & May only detect local optimums.                          &                           \\ \midrule
\hline
\multirow{2}{*}{BO-GP}                 & \multirow{2}{*}{Fast convergence for continuous HPs.} & Poor capacity for parallelization.                       & \multirow{2}{*}{$O(n^3)$} \\ \cmidrule(lr){3-3}
                                       &                                                             & Not efficient with conditional HPs.                      &                           \\ \midrule
\hline

\multirow{2}{*}{Hyperband}             & \multirow{2}{*}{Enable paralelliization}                    & Not efficient with conditional HPs.                      & $O(nlogn)$                \\ \cmidrule(l){3-3} 
                                       &                                                             & Require subsets with small budgets  &                                        \\ \midrule
\hline

\multirow{2}{*}{GA}                    & Efficient with all types of HPs                                      & \multirow{2}{*}{Poor capacity for parallelization.}                           & \multirow{2}{*}{$O(n^2)$}   \\ \cmidrule(lr){3-3}
                                       & Not require good initialization                                      &                            &                           \\ \midrule
\hline

\multirow{2}{*}{PSO}                    & Efficient with all types of HPs                                      & \multirow{2}{*}{Require proper initialization.}                           & \multirow{2}{*}{$O(nlogn)$}   \\ \cmidrule(lr){3-3}
                                       & Enable paralellization                                      &                           &                           \\ \midrule
\hline

\end{tabular}
\label{t1}
\end{table*}

\section{Quantum Machine Learning}\label{sec:QML}
Quantum machine learning (QML) \cite{schuld2015introduction, biamonte2017quantum, schuld2018supervised} explores the interplay and takes advantage of quantum computing and machine learning ideas and techniques.

Therefore, quantum machine learning is a hybrid system involving both classical and quantum processing, where computationally complex subroutines are given to quantum devices. QML tries to take advantage of the classical machine learning does best and what it costs, such as distance calculation (inner product), passing it onto a quantum computer that can compute it natively in the Hilbert vector space. 
In this era of large classical data and few qubits, the most common use is to design machine learning algorithms for classical data analysis running on a quantum computer, i.e., quantum-enhanced machine learning \cite{dunjko2016quantum, atchade2020using, atchade2022quantum, adelomou2022quantum, ParfaitVQE, Ket.G,schuld2018supervised}. 

\textit{Quantum circuits} are mathematically defined as operations on an initial quantum state. Quantum computing generally makes use of quantum states built from qubits, that is, binary states represented as  $\ket{\psi}=\alpha\ket{0} +\beta\ket{1}$. Their number of qubits $n$  commonly defines the states of a quantum circuit, and, in general, the circuit's initial state $\ket\psi_{0}$ is the zero state $\ket{0}$. In general, a quantum circuit implements an internal unit operation $U$  to the initial state $\ket\psi_{0}$ to transform it into the final output state $\ket\psi_{f}$. This gate $U$  is wholly fixed and known for some algorithms or problems. In contrast, others define its internal functioning through a fixed structure, called Ansatz\cite{Ansatz_best} (\textit{Parametrized Quantum Circuit} (PQC)), and adjustable parameters $\theta$ \cite{Suk191}. Parameterized circuits are beneficial and have interesting properties in this quantum age since they broadly define the definition of ML and provide flexibility and feasibility of unit operations with arbitrary precision \cite{JBi17,Adr20,Mar14}.

\begin{figure}[]
\centering
\includegraphics[width=0.45\textwidth]{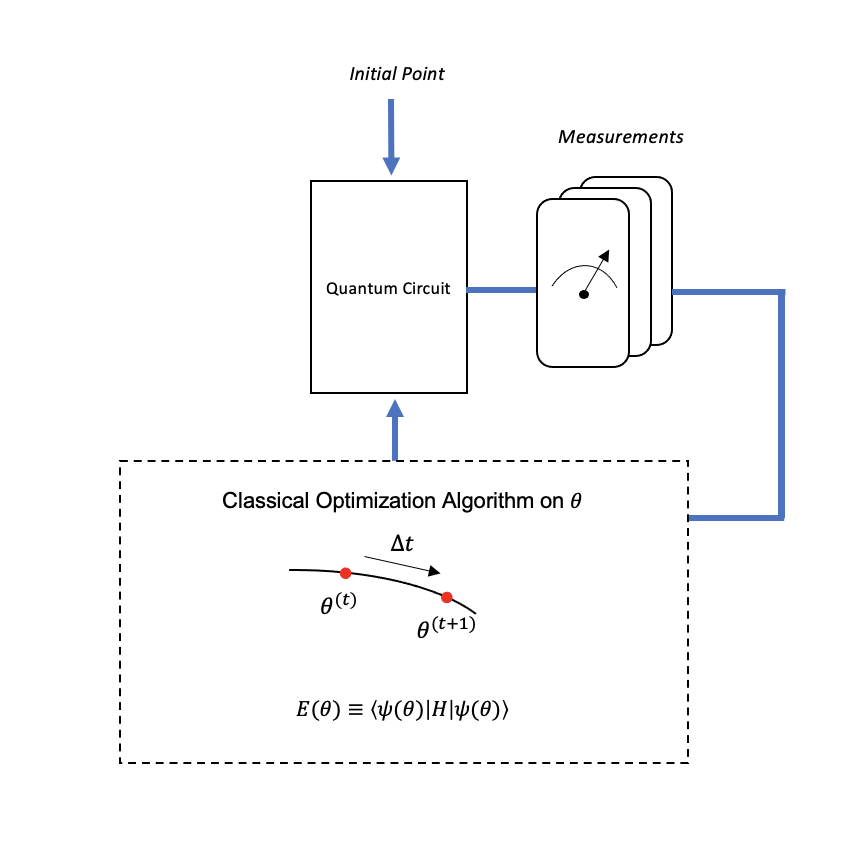}
\caption{\textit{Variational Quantum Algorithm} (VQC) working principle based on the quantum variational circuit. The quantum circuit computes the objective function, and the classical computer computes the circuit parameters. We can use this model to find the minima or the maxima of a given parameterized function that is our quantum circuit. Our quantum circuit will be separated into two large blocks: the \textit{Feature Map} and the \textit{Variational Quantum Circuit} 
 \cite{cerezo2021variational}.}
\label{fig:VQC}
\end{figure}

Figure \eqref{fig:VQC} depicts the concept of hybrid computing (quantum + classical), which characterizes the NISQ era. This takes advantage of quantum computing's capacity to solve complex problems and the experience of classical optimization algorithms (COBYLA\cite{The21}, SPSA\cite{Jam01}, BFGS\cite{BFGS_Limted}, etc.) to train variational circuits. Classical algorithms are generally an iterative scheme that searches for better candidates for the parameters $\theta$ at each step.

The value of the hybrid computing idea in the NISQ era is necessary because it allows the scientific community to exploit both capacities and reaps the benefits of the constant acceleration of the oncoming quantum-computer development.

Furthermore, learning techniques can be improved by embedding information (data) into the variational circuit through the quantum gate $U$ \cite{schuld2018supervised, perez2020data}.

The \textit{Variational Quantum Circuit} (VQC) \cite{Mic00, Mic} consists of a quantum circuit that defines the base structure similar to neural network architecture (Ansatz), while the variational procedure can optimize the types of gates (one or two-qubit parametric gates) and their free parameters.
The usual supervised learning processes within quantum machine learning can be defined as follows:

\begin{itemize}
    \item \textit{Quantum Feature Map}: It is the data preparation. In the literature, this stage is recognized as State preparation.
    \item \textit{The Quantum model}: It is the model creation. In the literature, it is recognized as unitary evolution.
    \item \textit{The classical error computation }: It is the stage of Computing the error where the model best approximates the input set; in machine learning, this stage is known as the prediction.
\end{itemize}

\section{IMPLEMENTATION}\label{sec:IMPLEMENT}
As aforementioned, our proposal is based on HPO.
\begin{figure*}[!ht]
\centering
\includegraphics[width=1\textwidth]{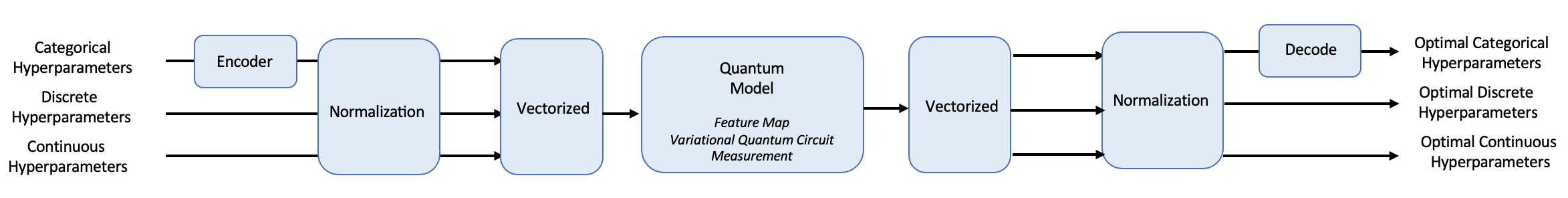}
\caption{This block diagram presents our approach of a generic hyperparameter tuning with quantum computing-based gradient descent or \textit{Adam optimizer} \cite{adamoptimizer}.
The first block allows us to normalize all inputs to $0$ and $\pi$. Then for the categorical variables, to save on the number of qubits, we encode the categorical variables in binary and map them over the number of qubits that our model has. The next block allows us to pack all the input variables into a single vector. Each variable is equivalent to a dimension of our vector. If the categorical variable is coded with $m$ dimensions, these $m$ dimensions will be directly mapped to the vector. Having the input vector, we now embed the data in the quantum computer thanks to our \textit{Feature Map} function. Said function will map our input variable continuously, with the help of our parameterized gates, $RX$, $RY$, and $RZ$. From here, we apply our quantum circuit. After the measurement, we must undo the previous operations in reverse order to recover our input hyperparameters.}
\label{fig:HPODomain}
\end{figure*}
Mathematically, we can formulate it as follows: Since a model's performance on a validation set can be modeled as a function $f: X \rightarrow \mathbb{R}$ of its hyperparameters $\vec{x} \in X$. With $X$ the hyperparameters' space and where $f$ can be any error function, such as the \textit{RMSE} in a regression problem or the \textit{AUC Score} for a classification problem. The problem that the HPO must solve is to find $\vec{x}$ such that $\vec{x} \in \text{argmin}_{\vec{x} \in X} f(\vec{x})$. 

Formally we can define our problem as follows. Let $f(\vec{x})$ be our objective function with $\vec{x}$ as the vector of all the classical input hyperparameters, and we are willing to find the best combination by writing it down as:

\begin{equation}
\vec{x}^{*} \approx \arg \min _{\vec{x} \in X} f(\vec{x}),
\label{eq:proposed_problem}
\end{equation}

In the optimization context, $f(x)$ is the objective function to be minimized, where $\vec{x}^{*}$ is the hyperparameter configuration that produces the optimum value of $f(\vec{x})$, and a hyperparameter $\vec{x}$ can take any value in the search space $X$.

The first step of our algorithm will be to find the function $f(\vec{x}) $ that generalizes our data. So, let us define our quantum model as follows: 
\begin{equation}
    f(\vec{x}) := \bra{0} U^{\dagger}(\vec{x},\beta, \vec{\theta}) \sigma_z U(\vec{x}, \beta, \vec{\theta}) \ket{0},
    \label{eq:model}
\end{equation}
Where $\sigma_z$ is our observable, $U(\vec{x}, \beta, \vec{\theta})$ the parameterized circuit with the input data $\Vec{x}$, $\beta$ our scaling factor and $\vec{\theta}$, the parameterized variable.

\begin{figure}[]
	\centering
		\includegraphics[width=0.45\textwidth]{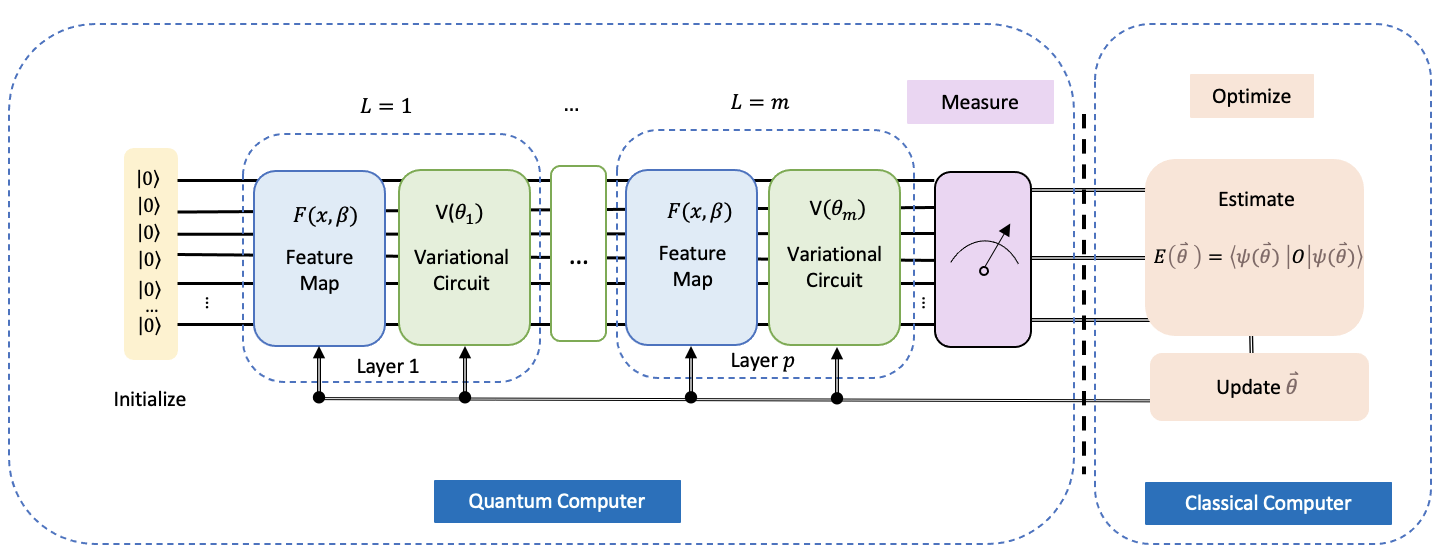}
		\caption{We use this model based on \textit{Fourier series} in quantum machine learning \cite{ParfaitAtchade}. Depending on some scaling $\beta$ parameters and the input data $\vec{x}$, the Feature Map will be in charge of coding our data. We rely heavily on the fact that the \textit{Feature Map} ($F(\vec{x},\beta)$) must be variational, that the data's loading is repeated in all the layers, and that the variational circuit ($V(\vec{\theta})$) searches with the help of the input parameters for the best function within the space of functions that defines the capacity of the variational circuit ($U(\vec{x},\beta,\vec{\theta})= F(\vec{x},\beta)V(\vec{\theta)}$). This configuration will efficiently approximate our given dataset to a continuous function $f(\vec{x})$.}
		\label{fig:qModel_four}
\end{figure}

From this point and considering equation \eqref{eq:proposed_problem}, we will find the minimum of \eqref{eq:model}.

In the next step, according to figure \eqref{fig:qproposal}, let $\beta$ and $\vec{\theta}$ be the parameters that define our function $f(\vec{x}) $ from equation \eqref{eq:model}. In this stage, we consider these parameters as constant inputs (we do not vary them). Now let $V_{\vec{\theta},\beta}(\vec{x})$ be the new circuit that can be an instantiation of the previous circuit ($U(\vec{x}, \beta, \vec{\theta})$), and let us execute the gradient of the said quantum circuit ($f(\vec{x}) $). The outcome of these operations will yield the optimal value of $f(\vec{x}) $, and the arguments for this operation will be the best hyperparameters we look for. The variable $\vec{x}$ will have the same dimension as the best hyperparameters we are finding.

From the \textit{variational principle} \cite{ekeland1974variational}, the following equation $\langle V_{\beta, \vec{\theta}}(\Vec{x}) \rangle _{ \psi   \left( \overrightarrow{ \gamma } \right) } \geq  \lambda _{i}$ can be reached. With $\lambda _{i}$  as eigenvector and  $\langle V_{\beta, \vec{\theta}}(\Vec{x}) \rangle _{ \psi \left( \overrightarrow{ \gamma } \right)}$  as the expected value. In this way, the VQC (Figure \eqref{fig:VQC}) finds \eqref{expectative_value} such an optimal choice of parameters $\overrightarrow{\gamma }$, that the expected value is minimized and that a lower eigenvalue is located. 
\begin{equation}
\label{expectative_value}
 \langle V_{\vec{\theta},\beta}(\Vec{x}) \rangle =\langle\psi\left(\gamma  \right)\vert V_{\vec{\theta},\beta}(\Vec{x}) \vert\psi\left(\gamma  \right)\rangle 
\end{equation}
Where $V_{\vec{\theta},\beta}(\Vec{x}) \approx f(\Vec{x}) $. 


\subsection{DATASET GENERATION}\label{sec:DBGENER}
The proposed quantum hyperparameter process is summarized in three stages. The first stage is the dataset generation, the second is finding the function that represents quantum-based Fourier regression, and the third is finding the minimum of the regression function. This section deals with dataset generation and criteria—the following subsection details all the experiment steps.
To implement the proposed algorithm, we generate one dataset per machine learning model regarding the original Vueling dataset considering its hyperparameters evaluating different search methods, as well as \textit{Grid search}, \textit{Random Search} and \textit{Bayes-Based Search}.

We train the $N$ ML models with the Vueling dataset to obtain a new reduced dataset. Three-way cross-validation is used during the training process. Therefore, for each set of hyperparameters, the model is trained three times and keeps the average of the three as the value to be predicted by the quantum system.

If more precision is needed, the generated dataset should have more elements. Instead of precision, whether the speed is prioritized at the expense of its precision, the number of features of the dataset should be considerably less than the number of features of the original dataset.

The resulting dataset size is reduced so as not to end up doing a quantum Grid Search. The new databases are stored in \textit{dataframes} using Pandas.

\subsection{EXPERIMENTS STEPS}\label{sec:EXPERIMENT}
\begin{figure}[]
	\centering
		\includegraphics[width=0.5\textwidth]{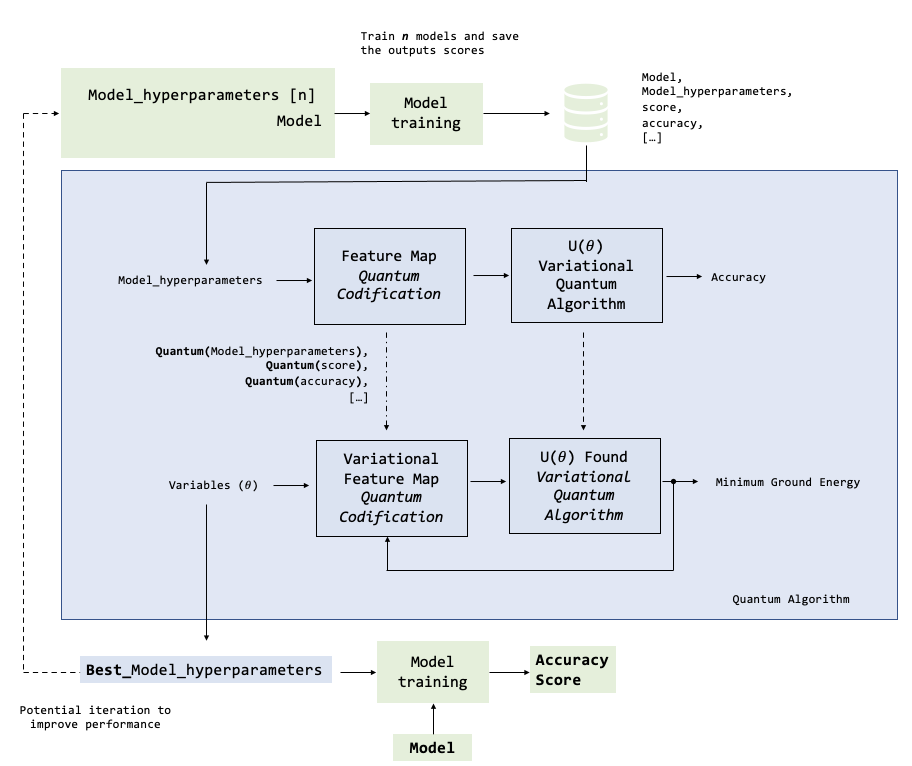}
		\caption{This graph shows the steps followed to achieve our goal. First, we generate a set ($[n]$) of randomly chosen hyperparameters and then train $n$ models with these hyperparameters. From the latter process, we get $[n]$ \textit{scores (accuracy, $r^2$, MSE, etc.)} and record the dataset. Next, we transform the categorical hyperparameter into linear variables according to figure \eqref{fig:HPODomain}. From this point, we transform this dataset into a Quantum Space for the native quantum algorithm. Having the dataset into the quantum domain, we now find the best $U(\vec{x},\beta,\vec{\theta})$ according to to figure \eqref{fig:qModel_four}. This means having our generic continuous function given by equation \eqref{eq:model}. In this stage, we only need to find the minimum, according to equation \eqref{eq:proposed_problem}. So, we train a Quantum-Coding System plus the best $U(\vec{x},\beta,\vec{\theta})$ found to predict the best hyperparameters. Now, we get the best set of hyperparameters by using a quantum computer (or a quantum-inspired one); we go back to the classical model and re-train it with the best hyperparameters.}
		\label{fig:qproposal}
\end{figure}

The scenario is given in Figure \eqref{fig:qproposal}, and the proposed process to validate our experimentation is summarized as follows: 
\begin{enumerate}
    \item From the database that contains $252,183$ datapoints with $42$ features each, 
    we generate a set ($[n]$) of randomly chosen hyperparameters. 
    \item We train $n$ models with these hyperparameters. Get $[n]$ \textit{scores (accuracy, $r^2$, MSE, etc.)} and  record the dataset.
    \item We transform the categorical hyperparameters into linear variables according to figure \eqref{fig:HPODomain}.
    \item We transform this dataset into a Quantum Space for the native quantum algorithm.
    \item We find the best $U(\vec{x},\beta,\vec{\theta})$ according to figure \eqref{fig:qModel_four}.
    \item We train a Quantum-Coding System plus the best $U(\vec{x},\beta,\vec{\theta})$ found to the predict best hyperparameters.
    \item We get the best set of hyperparameters by using Quantum Computer.
    \item We go back to the classical model and re-train it with the best hyperparameters.
\end{enumerate}

Classically there is a process $Classical\_process$ with its executing time $T_c$, and from quantum computing, there is a process $Quantum\_process$ with its executing time $T_q(translation) + T_q(solution)$. We will only get a time advantage if:
$T_c > [T_q(translation) + T_q(solution)]$.

Why $T_q(translation)?$ because quantum computers only execute quantum data, so we must translate all the data we got from the classical domain into the quantum one.

To test the proper functioning of our algorithm, we design five cases ($A$, $B$, $C$, $D$, and $E$), respectively, equivalent to the layer number of the variational algorithm ($1$, $2$, $3$, $4$, and $5$).

\subsection{OUR QUANTUM MODEL} \label{sec:modelQML}
Based on \cite{ParfaitAtchade}, we propose the hybrid model and strategy from figures \eqref{fig:HPODomain}, \eqref{fig:qModel_four} and \eqref{fig:qproposal} to tackle the generic HPO and reduce the search and analyzing time. 

\begin{table*}[]
\caption{This table represents the space's configuration for the hyperparameters and classical models that were tested in this work.}
\tiny
\begin{tabular}{l|l|l|l} 
\Xhline{1.2pt}
\textbf{ML Model }                       & \multicolumn{1}{l|}{\textbf{Hyperparameter}} & \multicolumn{1}{l|}{\textbf{Type}}        & \textbf{Search Space}                            \\ 
\Xhline{1.2pt}

\multirow{6}{*}{Random Forest Classifier}  & n\_estimators                        & Discrete                         & {[}5 - 250]                              \\ 
\cline{2-4}
                                & max\_depth                           & \multicolumn{1}{l|}{Discrete}    & {[}1 - 50]                                \\ 
\cline{2-4}
                                & min\_samples\_split                  & Discrete                         & {[}1 - 10]                                \\ 
\cline{2-4}
                                & min\_samples\_leaf                   & Discrete                         & {[}1 - 10]                                \\ 
\cline{2-4}
                                & criterion                            & \multicolumn{1}{l|}{Categorical} & {[}'gini', 'entropy', 'ROC']                   \\ 
\cline{2-4}
                                & max\_features                        & Discrete                         & {[}1 - 64]                                \\ 
\hline
\multirow{2}{*}{SVM Classifier} & C                                    & Continuous                       & {[}0.1,50]                              \\ 
\cline{2-4}
                                & kernel                               & Categorical                      & {[}'linear', 'poly', 'rbf', 'sigmoid']  \\ 
\hline
KNN Classifier                  & n\_neighbors                         & Discrete                         & {[}1,20]                                \\ 
\hline
\hline
\multirow{6}{*}{Random Forest Regressor}   & n\_estimators                        & Discrete                                                          & {[}5 - 250]                              \\ 
\cline{2-4}
                                & max\_depth                           & Discrete                         & {[}1 - 15]                                \\ 
\cline{2-4}
                                & min\_samples\_split                  & Discrete                         & {[}1 - 7]                                \\ 
\cline{2-4}
                                & min\_samples\_leaf                   & Discrete                         & {[}1 - 8]                                \\ 
\cline{2-3}\cline{4-4}
                                & criterion                            & Categorical                      & {[}'mse', 'mae' 'absolute\_error' 'friedman\_mse' ]                        \\ 

\hline
\multirow{3}{*}{SVM Regressor}  & max\_iter                                    & Continuous                                                & {[}1000 - 500000]                              \\ 
\cline{2-4}
                                & kernel                               & Categorical                      & {[}'linear', 'poly', 'rbf', 'sigmoid']  \\ 
\cline{2-4}
                                & epsilon                              & Continuous                       & {[}0.0001 - 1]                             \\ 
\hline
\multirow{3}{*}{HistGradientBoostingRegressor}  & max\_iter            & Discrete                         & {[}1 - 1000]                              \\ 
\cline{2-4}
                                & learning\_rate                       & Discreto                         & {[}0.01 - 1]   \\ 
                                
\cline{2-4}
                                & max\_bin                             & Discreto                         & {[}31 - 255]                             \\ 
\cline{2-4}
                                & loss                             & Categorical                         & {[} 'squared\_error' 'absolute\_error' 'poisson'  'quantile' ]                              \\                         
\hline

\multirow{3}{*}{Ridge}  & max\_iter                                      & Discrete                       &                          {[}1000 - 500000]                              \\ 
\cline{2-4}
                                & solver                               & Categorical                      & {[}'svd', 'cholesky', 'lsqr', 'sparse\_cq', 'sag' ]  \\ 
\cline{2-4}
                                & alpha                              & Continuous                       & {[}0.0001 - 1]                             \\ 
\hline
\multirow{3}{*}{Decision Tree Regressor}  & max\_depth                                      & Discrete                       &                          {[}1 - 100]                              \\ 
\cline{2-4}
                                & solver                               & Categorical                      & {[}'mse', 'mae' 'absolute\_error' 'friedman\_mse' 'poisson' ]  \\ 
\cline{2-4}
                                & ccp\_alpha                              & Continuous                       & {[}0.0 - 1]                             \\ 
\hline

\Xhline{1.2pt}
\end{tabular}
\label{ta1}%
\end{table*}

\section{Results} \label{sec:resultados}

\begin{figure}[]
	\centering
        \includegraphics[width=0.4\textwidth]{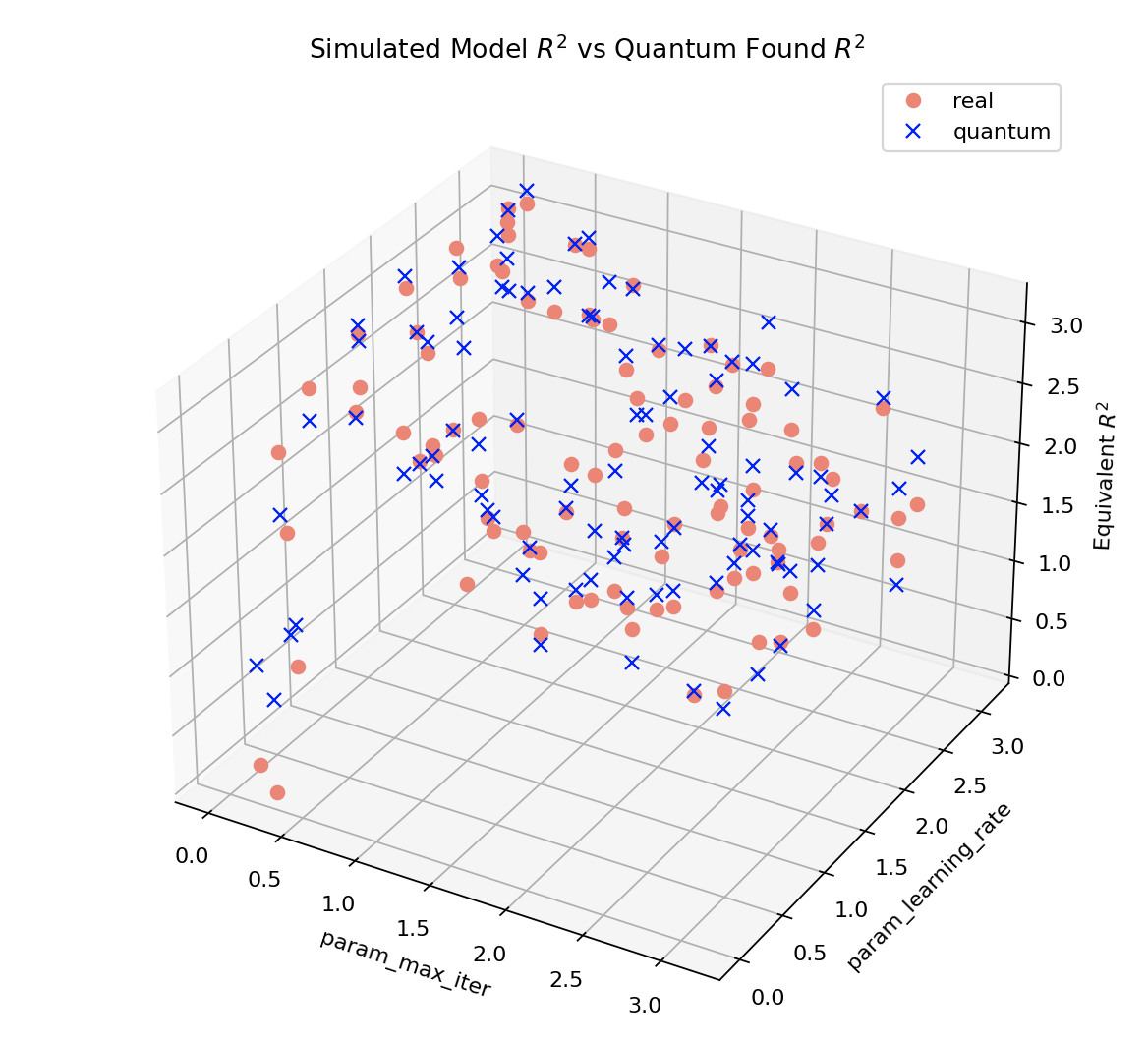}
		\caption{We can observe the result of applying a quantum-based Fourier regression approach. We can observe how the model allows interpolating the data from the dataset better.}
		\label{fig:resTrigoSquare}
\end{figure}

\begin{figure}[]
	\centering
        \includegraphics[width=0.4\textwidth]{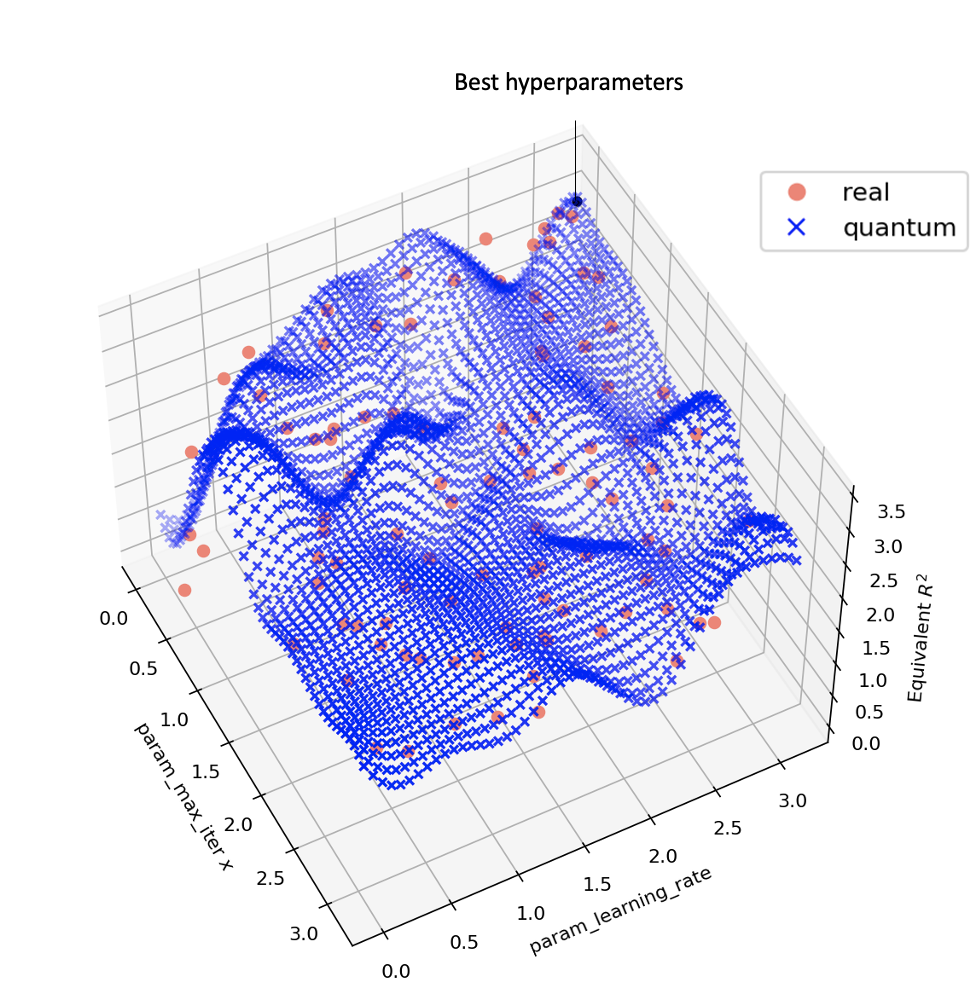}
		\caption{We can see the space covered by the new function $f(\vec{x})$ resulting from applying the gate $U(\vec{x},\vec{\theta}, \beta)$. In this case, the number of layers is $4$. The axes on the graph represent hyperparameters from \textit{HistGradBoost} model. Here, the score is normalized on $\pi$. Please refer to table \eqref{ta1} for more details.}
		\label{fig:resTrigoSquare_}
\end{figure}

\begin{figure}[b]
	\centering
        \includegraphics[width=0.4\textwidth]{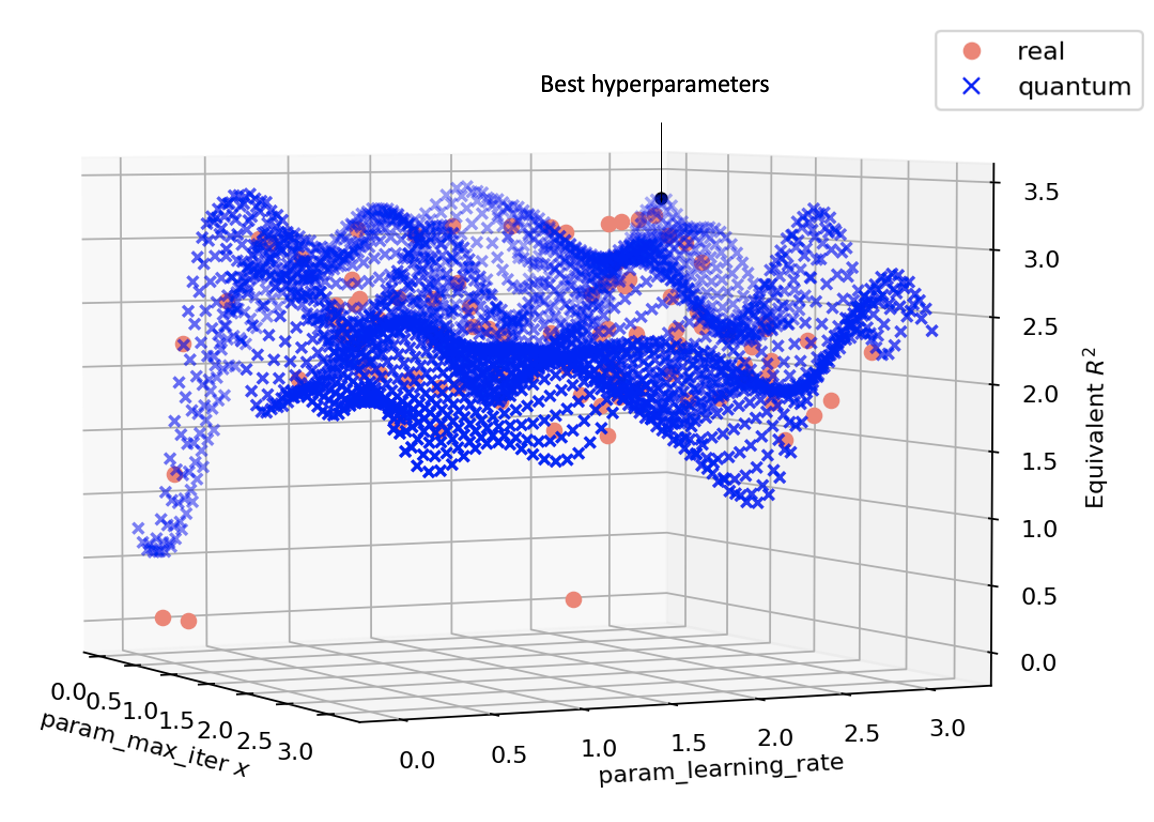}
		\caption{We can observe how our algorithm finds the local maximum of our multivariate function. In this case, the number of layers is $5$. The axes on the graph represent hyperparameters from \textit{HistGradBoost} model. Please refer to table \eqref{ta1} for more details.}
		\label{fig:resTrigoSquare__}
\end{figure}

Figures \eqref{fig:resTrigoSquare} to \eqref{fig:resTrigoSquare__} show the results of the process we follow to achieve our goal. It can be observed in Figure \eqref{fig:resTrigoSquare_} how the quantum model is molded into the shape of the data, defining a mesh that represents the continuous function, which is used to find the best hyperparameters.

We have tested the proposed algorithm and steps with various models and techniques and generated several databases and experiments. The detailed results of each experiment can be seen in the tables \eqref{resultsHistGradBoost_4547_LR_MaxIter_RandomCV}, \eqref{t_resultsHistGradBoost_GSCV4145}, \eqref{t_resultsRandomForest_GSCV}, and \eqref{Ridge_GCV}. In addition, we have carried out some comparative studies that can be seen in Figures \eqref{fig:benchmark_HistGradBoost_4547_Quantum} to \eqref{fig:benchmark_BayesSearchCV_Quantum}.

\subsection{2 Hyperparameters} \label{sec:2hyperparams}

From the database that contains $252,183$ datapoints with $42$ features each, we generate the database for the \textit{Histogram Gradient Boosting Regression}.
The result of our algorithm by using \textit{Histogram Gradient Boosting Regression}, precisely a \textit{lightGBM} implementation and executing the following tests will be shown in three hyperparameters' configurations ($2$, $3$, and $4$).

In the case of \textit{2 Hyperparameters}: \textit{learning\_rate and max\_iteration}, we obtain the following outcomes:

\begin{itemize}
    \item Testing classically with the full database with $240$ training samples took $6$ minutes $6$ seconds, with $R^2$ result for the $R^2_{\text{train}}= 0.464$ and $R^2_{\text{test}}= 0.388$.
    \item By creating a subset with only $70$ samples, that took $1$ minute $38$ seconds to be created and having the $R^2= 0.382$.
    \item Now, executing our quantum algorithm for finding the best hyperparameters using the prepared dataset ($70$ samples) took only  $4$ minutes and $43$ seconds to find the best hyperparameters  with $R^2$ score $= 0.444$.
\end{itemize}

\subsection{3 Hyperparameters} \label{sec:3hyperparams}
In the case of \textit{3 Hyperparameters}: \textit{learning\_rate, max\_iteration, and loss}, we obtain the following outcomes:

\begin{itemize}
    \item Testing classically with the full database with $480$ training samples took $11$ minutes $36$ seconds, with $R^2$ result for the $R^2_{\text{train}}= 0.464$ and $R^2_{\text{test}}= 0.388$.
    \item By creating a subset with only $35$ samples, that took $53$ seconds to be created and having the $R^2= 0.382$.
    \item Now, executing our quantum algorithm for finding the best hyperparameters using the prepared dataset ($35$ samples) took only  $1$ minute and $56$ seconds to find the best hyperparameters  with $R^2$ score $= 0.4377$.
\end{itemize}

\begin{table*}[t]
\tiny

\begin{tabular}{|l|rrrrr|}
\Xhline{1.2pt}
\textbf{Model}                                          & \multicolumn{5}{|c|}{\textbf{HistGradBoost with Random CV}}                                              \\ 
\Xhline{1.2pt}
\textbf{Cases}                                          & \multicolumn{1}{|c|}{A}       & \multicolumn{1}{c|}{B}       & \multicolumn{1}{c|}{C}       & \multicolumn{1}{c|}{D}       & \multicolumn{1}{c|}{E} \\ \hline
\textbf{Classical HistGradBoost performance time (s)}   & \multicolumn{1}{|r|}{2747.00} & \multicolumn{1}{r|}{2747.00} & \multicolumn{1}{r|}{2747.00} & \multicolumn{1}{r|}{2747.00} & 2747.00                \\ \hline
\textbf{Total proposed model  performance time (s)}     & \multicolumn{1}{|r|}{146.86}  & \multicolumn{1}{r|}{206.09}  & \multicolumn{1}{r|}{313.46}  & \multicolumn{1}{r|}{374.91}  & 555.11                 \\ \hline
\textbf{Time saving (s)}                                & \multicolumn{1}{|r|}{2598.14} & \multicolumn{1}{r|}{2538.91} & \multicolumn{1}{r|}{2431.54} & \multicolumn{1}{r|}{2370.09} & 2189.89                \\ \hline
\textbf{Time saving (\%)}                               & \multicolumn{1}{|r|}{94.65} & \multicolumn{1}{r|}{92.49} & \multicolumn{1}{r|}{89.59} & \multicolumn{1}{r|}{86.35} & 79.79                \\ \hline
\textbf{Dev Score}                                      & \multicolumn{1}{|r|}{0.00}    & \multicolumn{1}{r|}{0.00}    & \multicolumn{1}{r|}{0.00}    & \multicolumn{1}{r|}{0.00}    & 0.00                   \\ \hline
\textbf{Dev Score (\%)}                                 & \multicolumn{1}{|r|}{0.00}    & \multicolumn{1}{r|}{0.00}    & \multicolumn{1}{r|}{0.00}    & \multicolumn{1}{r|}{0.00}    & 0.00                   \\ \hline
\textbf{\# HPs}                                         & \multicolumn{1}{|r|}{3.00}    & \multicolumn{1}{r|}{3.00}    & \multicolumn{1}{r|}{3.00}    & \multicolumn{1}{r|}{3.00}    & 5.00                   \\ \hline
\textbf{\# Layers}                                      & \multicolumn{1}{|r|}{1.00}    & \multicolumn{1}{r|}{2.00}    & \multicolumn{1}{r|}{3.00}    & \multicolumn{1}{r|}{4.00}    & 5.00                   \\ \hline
\textbf{Load Data(ms)}                                   & \multicolumn{1}{|r|}{1.30}    & \multicolumn{1}{r|}{2.40}    & \multicolumn{1}{r|}{4.10}    & \multicolumn{1}{r|}{4.81}    & 0.00                   \\ \hline
\textbf{VQA (s)}                                        & \multicolumn{1}{|r|}{84.70}   & \multicolumn{1}{r|}{150.05}  & \multicolumn{1}{r|}{239.06}  & \multicolumn{1}{r|}{282.38}  & 397.03                 \\ \hline
\textbf{Finding Quantum best HPs (s)}                   & \multicolumn{1}{|r|}{28.63}   & \multicolumn{1}{r|}{44.63}   & \multicolumn{1}{r|}{68.01}   & \multicolumn{1}{r|}{85.49}   & 140.42                 \\ \hline
\textbf{Data mapping from Quantum to Classic Space ($\mu$s)} & \multicolumn{1}{|r|}{30.35}    & \multicolumn{1}{r|}{41.76}    & \multicolumn{1}{r|}{51.51}    & \multicolumn{1}{r|}{70.66}    & 73.78                   \\ \hline
\textbf{Loading Original dataset time (s)}              & \multicolumn{1}{|r|}{1.44}    & \multicolumn{1}{r|}{1.07}    & \multicolumn{1}{r|}{1.10}    & \multicolumn{1}{r|}{1.07}    & 1.09                   \\ \hline
\textbf{Model training (s)}                             & \multicolumn{1}{|r|}{32.08}   & \multicolumn{1}{r|}{10.34}   & \multicolumn{1}{r|}{5.29}    & \multicolumn{1}{r|}{5.97}    & 16.56                  \\ \hline
\textbf{Proposed model Test score}                             & \multicolumn{1}{|r|}{0.3879}    & \multicolumn{1}{r|}{0.39}    & \multicolumn{1}{r|}{0.3879}    & \multicolumn{1}{r|}{0.3879}    & 0.3879                   \\ \hline
\textbf{Orginal Train score}                            & \multicolumn{1}{|r|}{0.4578}    & \multicolumn{1}{r|}{0.4598}    & \multicolumn{1}{r|}{0.4578}    & \multicolumn{1}{r|}{0.4578}    & 0.4598                   \\ \hline
\textbf{Orginal Test Score}                             & \multicolumn{1}{|r|}{0.3879}    & \multicolumn{1}{r|}{0.3879}    & \multicolumn{1}{r|}{0.3879}    & \multicolumn{1}{r|}{0.3879}    & 0.3879                   \\ 
\Xhline{1.2pt}
\end{tabular}
\caption{
    This table shows the experiments for five cases (A, B, C, D, and E) with the input database trained with both the test and train for the \textit{HistGradBoost} model. The time required to generate said database classically and have the optimal hyperparameters applying \textit{Random Search} with \textit{Cross-Validation} is $2747$ seconds. The configuration parameters of our hybrid model to obtain the data from the tables are the following: $\textit{lrVQA}=0.15$, $\textit{maxEpochVQA}=70$, $\textit{lrBH}=0.0005$, $\textit{maxEpochBH}=1500$, $loadOptBH=False$ and for the different quantum layers (\textit{qLayer}$=1, 2, 3, 4,$ \text{and}  $5$). Our algorithm already finds the hyperparameters for the defined target score, considerably reducing the experimentation time using minimum layers. In this case, from $94\%$ for one layer to $79\%$ for five layers. All the tests were done locally on \textit{Mac Book Pro} with \textit{8-Core Intel} \cite{MacBook}.
}
\label{resultsHistGradBoost_4547_LR_MaxIter_RandomCV}
\end{table*}

\subsection{4 Hyperparameters} \label{sec:4hyperparams}

In the case of  \textit{4 Hyperparameters}: \textit{learning\_rate, max\_iteration, loss and max\_bins}, we have the following outcomes:
\begin{itemize}
    \item Testing classically with the full database with $1980$ training samples took $41$ minutes $45$ seconds, with $R^2$ result for the $R^2_{\text{train}}= 0.464$ and $R^2_{\text{test}}= 0.388$.
    \item By creating a subset with only $280$ samples, that took $5$ minutes $54$ seconds to be created and having the $R^2= 0.388$.
    \item Now, executing our quantum algorithm for finding the best hyperparameters using the prepared dataset ($280$ samples) took only  $15$ minutes and $2$ seconds to find the best hyperparameters  with $R^2$ score $= 0.415$.
\end{itemize}


\begin{table*}[]
\tiny
\begin{tabular}{|l|rrrrr|}
\Xhline{1.2pt}
\textbf{Model}                                          & \multicolumn{5}{|c|}{\textbf{HistGradBoost with Grid CV}}                                \\ 
\Xhline{1.2pt}
\textbf{Cases}                                          & \multicolumn{1}{|c|}{A}       & \multicolumn{1}{c|}{B}       & \multicolumn{1}{c|}{C}       & \multicolumn{1}{c|}{D}       & \multicolumn{1}{c|}{E} \\ \hline
\textbf{Classical HistGradBoost performance time (s)}   & \multicolumn{1}{|r|}{2505.00} & \multicolumn{1}{r|}{2505.00} & \multicolumn{1}{r|}{2505.00} & \multicolumn{1}{r|}{2505.00} & 2505.00                \\ \hline
\textbf{Total proposed model  performance time (s)}     & \multicolumn{1}{|r|}{368.30}  & \multicolumn{1}{r|}{774.53} & \multicolumn{1}{r|}{944.76}  & \multicolumn{1}{r|}{1470.23}  & 1757.00                \\ \hline
\textbf{Time saving (s)}                                & \multicolumn{1}{|r|}{2136.69} & \multicolumn{1}{r|}{1730.46} & \multicolumn{1}{r|}{1560.24} & \multicolumn{1}{r|}{1034.76} & 747.9930                \\ \hline
\textbf{Time saving (\%)}                               & \multicolumn{1}{|r|}{85.29} & \multicolumn{1}{r|}{69.08} & \multicolumn{1}{r|}{62.28} & \multicolumn{1}{r|}{41.31} & 29.86                \\ \hline
\textbf{Dev Score}                                      & \multicolumn{1}{|r|}{0.01}    & \multicolumn{1}{r|}{0.01}    & \multicolumn{1}{r|}{0.01}    & \multicolumn{1}{r|}{0.01}    & 0.01                   \\ \hline
\textbf{Dev Score (\%)}                                 & \multicolumn{1}{|r|}{0.01}    & \multicolumn{1}{r|}{0.01}    & \multicolumn{1}{r|}{0.01}    & \multicolumn{1}{r|}{0.01}    & 0.01                   \\ \hline
\textbf{\# HPs}                                         & \multicolumn{1}{|r|}{4.00}    & \multicolumn{1}{r|}{4.00}    & \multicolumn{1}{r|}{4.00}    & \multicolumn{1}{r|}{4.00}    & 4.00                   \\ \hline
\textbf{\# Layers}                                      & \multicolumn{1}{|r|}{1.00}    & \multicolumn{1}{r|}{2.00}    & \multicolumn{1}{r|}{3.00}    & \multicolumn{1}{r|}{4.00}    & 5.00                   \\ \hline
\textbf{Load Data(s)}                                   & \multicolumn{1}{|r|}{0.00}    & \multicolumn{1}{r|}{0.00}    & \multicolumn{1}{r|}{0.00}    & \multicolumn{1}{r|}{0.00}    & 0.00                   \\ \hline
\textbf{VQA (s)}                                        & \multicolumn{1}{|r|}{308.31}  & \multicolumn{1}{r|}{700.37}  & \multicolumn{1}{r|}{855.79}  & \multicolumn{1}{r|}{1302.36} & 1585.81                \\ \hline
\textbf{Finding Quantum best HPs (s)}                   & \multicolumn{1}{|r|}{45.68}   & \multicolumn{1}{r|}{65.80}   & \multicolumn{1}{r|}{81.75}   & \multicolumn{1}{r|}{146.56}  & 160.69                 \\ \hline
\textbf{Data mapping from Quantum to Classic Space (s)} & \multicolumn{1}{|r|}{0.00}    & \multicolumn{1}{r|}{0.00}    & \multicolumn{1}{r|}{0.00}    & \multicolumn{1}{r|}{0.00}    & 0.00                   \\ \hline
\textbf{Loading Original dataset time (s)}              & \multicolumn{1}{|r|}{1.60}    & \multicolumn{1}{r|}{1.11}    & \multicolumn{1}{r|}{1.00}    & \multicolumn{1}{r|}{1.60}    & 1.23                   \\ \hline
\textbf{Model training (s)}                             & \multicolumn{1}{|r|}{12.72}   & \multicolumn{1}{r|}{7.25}    & \multicolumn{1}{r|}{6.21}    & \multicolumn{1}{r|}{19.72}   & 9.27                   \\ \hline
\textbf{Proposed model Test score}                             & \multicolumn{1}{|r|}{0.3785}    & \multicolumn{1}{r|}{0.3785}    & \multicolumn{1}{r|}{0.3785}    & \multicolumn{1}{r|}{0.3785}    & 0.3785                   \\ \hline
\textbf{Original Train score}                            & \multicolumn{1}{|r|}{0.4583}    & \multicolumn{1}{r|}{0.4583}    & \multicolumn{1}{r|}{0.4583}    & \multicolumn{1}{r|}{0.4583}    & 0.4583                   \\ \hline
\textbf{Original Test Score}                             & \multicolumn{1}{|r|}{0.3879}    & \multicolumn{1}{r|}{0.3879}    & \multicolumn{1}{r|}{0.3879}    & \multicolumn{1}{r|}{0.3879}    & 0.3879                   \\ 
\Xhline{1.2pt}
\end{tabular}
\caption{
This table shows the experiments for five cases (A, B, C, D, and E) with the input database trained with both the test and train for the \textit{HistGradBoost} model. The time required to generate said database classically and have the optimal hyperparameters applying \textit{Grid Search} with \textit{Cross-Validation} is $2505$ seconds. The configuration parameters of our hybrid model to obtain the data from the tables are the following: $\textit{lrVQA}=0.15$, $\textit{maxEpochVQA}=70$, $\textit{lrBH}=0.0005$, $\textit{maxEpochBH}=1500$, $loadOptBH=False$ and for the different quantum layers (\textit{qLayer}$=1, 2, 3, 4,$ \text{and}  $5$). Our algorithm already finds the hyperparameters for the defined target score, considerably reducing the experimentation time using minimum layers. In this case, from $85\%$ for one layer to $29\%$ for five layers. All the tests were done locally on \textit{MacBookPro} with \textit{8-Core Intel} \cite{MacBook}.
}
\label{t_resultsHistGradBoost_GSCV4145}
\end{table*}

\begin{table*}[t]
\tiny
\begin{tabular}{|l|rrrrr|}
\Xhline{1.2pt}
\textbf{Model}                                          & \multicolumn{5}{|c|}{\textbf{RandomForest with Grid CV}}                                       \\ 
\Xhline{1.2pt}
\textbf{Cases}                                          & \multicolumn{1}{|c|}{A}       & \multicolumn{1}{c|}{B}       & \multicolumn{1}{c|}{C}       & \multicolumn{1}{c|}{D}       & \multicolumn{1}{c|}{E} \\ \hline
\textbf{Classical Random Forest performance time (s)}   & \multicolumn{1}{|r|}{2116.00} & \multicolumn{1}{r|}{2116.00} & \multicolumn{1}{r|}{2116.00} & \multicolumn{1}{r|}{2116.00} & 2116.00                \\ \hline
\textbf{Total proposed model  performance time (s)}     & \multicolumn{1}{|r|}{669.22}  & \multicolumn{1}{r|}{1018.24} & \multicolumn{1}{r|}{932.06}  & \multicolumn{1}{r|}{955.51}  & 1021.88                \\ \hline
\textbf{Time saving (s)}                                & \multicolumn{1}{|r|}{1446.77} & \multicolumn{1}{r|}{1097.76} & \multicolumn{1}{r|}{1183.93} & \multicolumn{1}{r|}{1160.48} & 1094.12                \\ \hline
\textbf{Time saving (\%)}                               & \multicolumn{1}{|r|}{68.37} & \multicolumn{1}{r|}{51.88} & \multicolumn{1}{r|}{55.95} & \multicolumn{1}{r|}{54.84} & 51.71                \\ \hline
\textbf{Dev Score}                                      & \multicolumn{1}{|r|}{0.00}    & \multicolumn{1}{r|}{0.00}    & \multicolumn{1}{r|}{0.00}    & \multicolumn{1}{r|}{0.00}    & 0.00                   \\ \hline
\textbf{Dev Score (\%)}                                 & \multicolumn{1}{|r|}{0.00}    & \multicolumn{1}{r|}{0.00}    & \multicolumn{1}{r|}{0.00}    & \multicolumn{1}{r|}{0.00}    & 0.00                   \\ \hline
\textbf{\# HPs}                                         & \multicolumn{1}{|r|}{2.00}    & \multicolumn{1}{r|}{2.00}    & \multicolumn{1}{r|}{2.00}    & \multicolumn{1}{r|}{2.00}    & 2.00                   \\ \hline
\textbf{\# Layers}                                      & \multicolumn{1}{|r|}{1.00}    & \multicolumn{1}{r|}{2.00}    & \multicolumn{1}{r|}{3.00}    & \multicolumn{1}{r|}{4.00}    & 5.00                   \\ \hline
\textbf{Load Data(s)}                                   & \multicolumn{1}{|r|}{0.00}    & \multicolumn{1}{r|}{0.00}    & \multicolumn{1}{r|}{0.00}    & \multicolumn{1}{r|}{0.00}    & 0.00                   \\ \hline
\textbf{VQA (s)}                                        & \multicolumn{1}{|r|}{80.04}   & \multicolumn{1}{r|}{101.80}  & \multicolumn{1}{r|}{148.99}  & \multicolumn{1}{r|}{218.43}  & 245.46                 \\ \hline
\textbf{Finding Quantum best HPs (s)}                   & \multicolumn{1}{|r|}{23.35}   & \multicolumn{1}{r|}{37.88}   & \multicolumn{1}{r|}{53.02}   & \multicolumn{1}{r|}{70.19}   & 76.26                  \\ \hline
\textbf{Data mapping from Quantum to Classic Space (s)} & \multicolumn{1}{|r|}{0.00}    & \multicolumn{1}{r|}{0.00}    & \multicolumn{1}{r|}{0.00}    & \multicolumn{1}{r|}{0.00}    & 0.00                   \\ \hline
\textbf{Loading Original dataset time (s)}              & \multicolumn{1}{|r|}{1.29}    & \multicolumn{1}{r|}{1.49}    & \multicolumn{1}{r|}{1.18}    & \multicolumn{1}{r|}{1.30}    & 1.02                   \\ \hline
\textbf{Model training (s)}                             & \multicolumn{1}{|r|}{564.55}    & \multicolumn{1}{r|}{877.07}  & \multicolumn{1}{r|}{728.88}  & \multicolumn{1}{r|}{665.60}  & 699.14                 \\ \hline
\textbf{Proposed model Test score}                             & \multicolumn{1}{|r|}{0.3478}    & \multicolumn{1}{r|}{3478}    & \multicolumn{1}{r|}{0.3478}    & \multicolumn{1}{r|}{0.3478}    & 0.3478                   \\ \hline
\textbf{Original Train score}                            & \multicolumn{1}{|r|}{0.6273}    & \multicolumn{1}{r|}{0.6273}    & \multicolumn{1}{r|}{0.6273}    & \multicolumn{1}{r|}{0.6273}    & 0.6273                   \\ \hline
\textbf{Original Test Score}                             & \multicolumn{1}{|r|}{0.3478}    & \multicolumn{1}{r|}{0.3478}    & \multicolumn{1}{r|}{0.3478}    & \multicolumn{1}{r|}{0.3478}    & 0.3478                   \\ 
\Xhline{1.2pt}
\end{tabular}
\caption{
This table shows the experiments for five cases (A, B, C, D, and E) with the input database trained with both the test and train for the \textit{RandomForest} model. The time required to generate said database classically and have the optimal hyperparameters applying \textit{Grid Search} with \textit{Cross-Validation} is $2116$ seconds. The configuration parameters of our hybrid model to obtain the data from the tables are the following: $\textit{lrVQA}=0.15$, $\textit{maxEpochVQA}=70$, $\textit{lrBH}=0.0005$, $\textit{maxEpochBH}=1500$, $loadOptBH=False$ and for the different quantum layers (\textit{qLayer}$=1, 2, 3, 4,$ \text{and}  $5$). Our algorithm already finds the hyperparameters for the defined target score, considerably reducing the experimentation time using minimum layers. In this case, from $68\%$ for one layer to $51\%$ for five layers. All the tests were done locally on \textit{MacBookPro} with \textit{8-Core Intel} \cite{MacBook}.
}
\label{t_resultsRandomForest_GSCV}
\end{table*}

\begin{table*}[t]
\tiny
\begin{tabular}{|l|rrrrr|}
\Xhline{1.2pt}
\textbf{Model}                                          & \multicolumn{5}{|c|}{\textbf{Ridge with Grid Search CV}}                                            \\ \Xhline{1.2pt}
\textbf{Cases}                                          & \multicolumn{1}{|c|}{A}       & \multicolumn{1}{c|}{B}       & \multicolumn{1}{c|}{C}       & \multicolumn{1}{c|}{D}       & \multicolumn{1}{c|}{E} \\ \hline
\textbf{Classical Ridge performance time (s)}           & \multicolumn{1}{|r|}{2075.00} & \multicolumn{1}{r|}{2075.00} & \multicolumn{1}{r|}{2075.00} & \multicolumn{1}{r|}{2075.00} & 2075.00                \\ \hline
\textbf{Total proposed model  time (s)}                 & \multicolumn{1}{|r|}{163.13}  & \multicolumn{1}{r|}{314.93}  & \multicolumn{1}{r|}{538.35}  & \multicolumn{1}{r|}{632.64}  & 871.59                 \\ \hline
\textbf{Time saving (s)}                                & \multicolumn{1}{|r|}{1911.87} & \multicolumn{1}{r|}{1760.07} & \multicolumn{1}{r|}{1536.65} & \multicolumn{1}{r|}{1442.36} & 1203.41                \\ \hline
\textbf{Time saving (\%)}                               & \multicolumn{1}{|r|}{92.13} & \multicolumn{1}{r|}{84.82} & \multicolumn{1}{r|}{74.05} & \multicolumn{1}{r|}{69.51} & 57.99                \\ \hline
\textbf{Dev Score}                                      & \multicolumn{1}{|r|}{0.00}    & \multicolumn{1}{r|}{0.00}    & \multicolumn{1}{r|}{0.00}    & \multicolumn{1}{r|}{0.00}    & 0.00                   \\ \hline
\textbf{Dev Score (\%)}                                 & \multicolumn{1}{|r|}{0.00}    & \multicolumn{1}{r|}{0.00}    & \multicolumn{1}{r|}{0.00}    & \multicolumn{1}{r|}{0.00}    & 0.00                   \\ \hline
\textbf{\# HPs}                                         & \multicolumn{1}{|r|}{5.00}    & \multicolumn{1}{r|}{5.00}    & \multicolumn{1}{r|}{5.00}    & \multicolumn{1}{r|}{5.00}    & 5.00                   \\ \hline
\textbf{\# Layers}                                      & \multicolumn{1}{|r|}{1.00}    & \multicolumn{1}{r|}{2.00}    & \multicolumn{1}{r|}{3.00}    & \multicolumn{1}{r|}{4.00}    & 5.00                   \\ \hline
\textbf{Load Data(s)}                                   & \multicolumn{1}{|r|}{0.00}    & \multicolumn{1}{r|}{0.00}    & \multicolumn{1}{r|}{0.00}    & \multicolumn{1}{r|}{0.00}    & 0.00                   \\ \hline
\textbf{VQA (s)}                                        & \multicolumn{1}{|r|}{118.77}  & \multicolumn{1}{r|}{243.11}  & \multicolumn{1}{r|}{403.12}  & \multicolumn{1}{r|}{490.68}  & 678.17                 \\ \hline
\textbf{Finding Quantum best HPs (s)}                   & \multicolumn{1}{|r|}{39.87}   & \multicolumn{1}{r|}{67.79}   & \multicolumn{1}{r|}{130.66}  & \multicolumn{1}{r|}{137.81}  & 188.86                 \\ \hline
\textbf{Data mapping from Quantum to Classic Space (s)} & \multicolumn{1}{|r|}{0.00}    & \multicolumn{1}{r|}{0.00}    & \multicolumn{1}{r|}{0.00}    & \multicolumn{1}{r|}{0.00}    & 0.00                   \\ \hline
\textbf{Loading Original dataset time (s)}              & \multicolumn{1}{|r|}{1.15}    & \multicolumn{1}{r|}{0.98}    & \multicolumn{1}{r|}{1.10}    & \multicolumn{1}{r|}{1.07}    & 1.13                   \\ \hline
\textbf{Model training (s)}                             & \multicolumn{1}{|r|}{3.34}    & \multicolumn{1}{r|}{3.04}    & \multicolumn{1}{r|}{3.46}    & \multicolumn{1}{r|}{3.07}    & 3.44                   \\ \hline
\textbf{Proposed model Test score}                             & \multicolumn{1}{|r|}{0.3012}    & \multicolumn{1}{r|}{0.3012}    & \multicolumn{1}{r|}{0.3012}    & \multicolumn{1}{r|}{0.3012}    & 0.3012                   \\ \hline
\textbf{Original Train score}                            & \multicolumn{1}{|r|}{0.3134}    & \multicolumn{1}{r|}{0.3134}    & \multicolumn{1}{r|}{0.3134}    & \multicolumn{1}{r|}{0.3134}    & 0.3134                   \\ \hline
\textbf{Original Test Score}                             & \multicolumn{1}{|r|}{0.3012}    & \multicolumn{1}{r|}{0.3012}    & \multicolumn{1}{r|}{0.3012}    & \multicolumn{1}{r|}{0.3012}    & 0.3012                   \\ 
\Xhline{1.2pt}

\end{tabular}
\caption{
This table shows the experiments for five cases (A, B, C, D, and E) with the input database trained with both the test and train for the \textit{Ridge} model. The time required to generate said database classically and have the optimal hyperparameters applying \textit{Grid Search} with \textit{Cross-Validation} is $2116$ seconds. The configuration parameters of our hybrid model to obtain the data from the tables are the following: $\textit{lrVQA}=0.15$, $\textit{maxEpochVQA}=70$, $\textit{lrBH}=0.0005$, $\textit{maxEpochBH}=1500$, $loadOptBH=False$ and for the different quantum layers (\textit{qLayer}$=1, 2, 3, 4,$ \text{and}  $5$). Our algorithm already finds the hyperparameters for the defined target score, considerably reducing the experimentation time using minimum layers. In this case, from $92\%$ for one layer to $57\%$ for five layers. All the tests were done locally on \textit{MacBookPro} with \textit{8-Core Intel} \cite{MacBook}.
}
\label{Ridge_GCV}
\end{table*}

\subsection{Discussions}\label{sec:discussions}

In this Section, we discuss the obtained results while elaborating on which cases our proposed model works best. 

\begin{figure}[]
	\centering
        \includegraphics[width=0.45\textwidth]{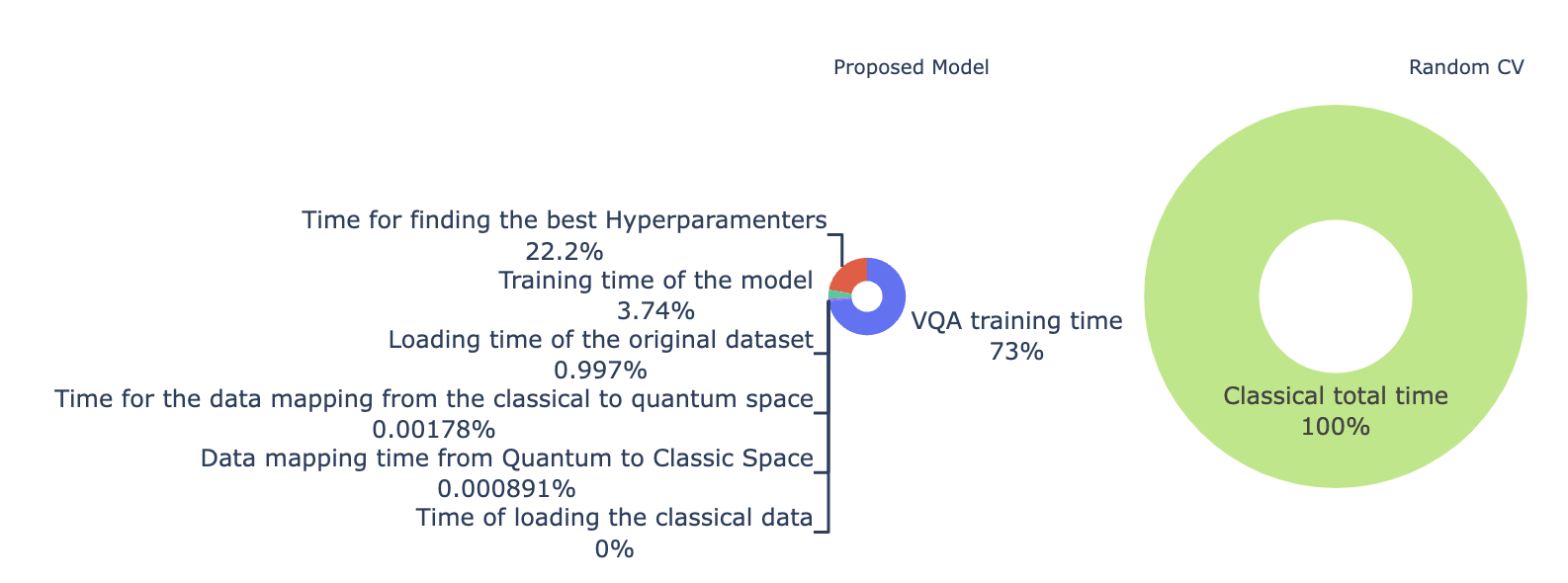}
		\caption{This graph compares the classical model and the hybrid proposed model—the classic model on the right of the image and the proposed model on the left. We are comparing the classically trained data model with the \textit{HistGradBoost} model using \textit{Random Search} with \textit{Cross-Validation}. The ratio between the graphs results from saving time in this case, with only a single layer in the quantum circuit. Table \eqref{resultsHistGradBoost_4547_LR_MaxIter_RandomCV} analyzes in detail considering the number of layers. In this case, we save $56\%$ of time compared to the classical counterpart.  A $1500$-element database with $3$ columns considering the following the \textit{hyperparameters param\_learning\_rate} $[0.01 - 1]$, \textit{param\_iter} $[1 - 1000]$, and \textit{param\_loss} as categorical $[$squared\_error, absolute\_error$]$ (for more details \eqref{ta1}).}
		\label{fig:benchmark_HistGradBoost_4547_Quantum}
\end{figure}

We compared the classical and hybrid proposed models from figure \eqref{fig:benchmark_HistGradBoost_4547_Quantum}. Specifically, we compared the classically trained data model with the \textit{HistGradBoost} model using \textit{Random Search} and \textit{Cross-Validation} with our model. The ratio between the graphs is from saving time, with only a single layer in the quantum circuit. 
In the case of figure \eqref{fig:benchmark_HistGradBoost_4547_Quantum}, we notice that we saved $56\%$ of total time.  The VQA took about $100$ seconds, the search algorithm for the optimal hyperparameters took about $33$ seconds, and the model retraining took about $50$ seconds compared to $2747$ seconds to get the same hyperparameters. Using Random search, the experiment was done with a $1500$-element database over $3$ columns. With the hyperparameters \textit{param\_learning\_rate} $[0.01 - 1]$, \textit{param\_iter} $[1 - 1000]$, and \textit{param\_loss} as categorical $[$squared\_error, absolute\_error$]$ (for more details \eqref{ta1}).

\begin{figure}[]
	\centering
        \includegraphics[width=0.45\textwidth]{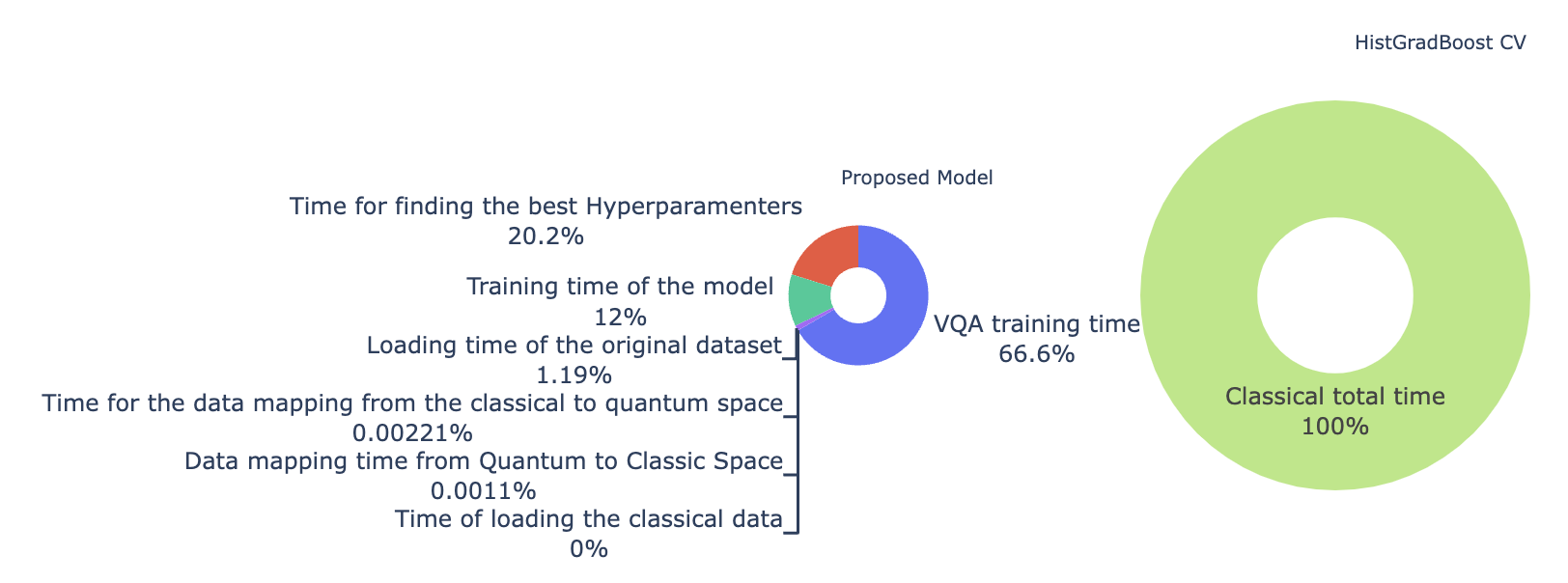}
		\caption{This graph compares the classical model and our hybrid proposal—the classic model on the right of the image and the proposed model on the left. Specifically, we are comparing the classically trained data model with the \textit{HisGradBoost} model using \textit{Grid Search} and \textit{Cross-Validation} with the model. The ratio between the graphs results from saving time in this case, with only a single layer in the quantum circuit. Table \eqref{t_resultsHistGradBoost_GSCV4145} analyzes in detail considering the number of layers. In this case, we save $81\%$ of time compared to the classical counterpart. A $1920$-element database over $4$ columns considering the following hyperparameters \textit{param\_learning\_rate} $[0.01 - 1]$, \textit{param\_iter} $[1 - 1000]$, \textit{param\_max\_bins} $[31 - 255]$ and \textit{param\_loss} as categorical $[$squared\_error, absolute\_error$]$ (for more details \eqref{ta1}).}
		\label{fig:benchmark_t_resultsHistGradBoost_GSCV4145_Quantum}
\end{figure}

Figure \eqref{fig:benchmark_t_resultsHistGradBoost_GSCV4145_Quantum} shows the experiment with the same model (\textit{HisGradBoost}) but this time with \textit{Grid Search} and \textit{Cross-Validation}. In this case, the VQA took about $425$ seconds, the search algorithm for the optimal hyperparameters took about $36$ seconds, and the model retraining took about $8$ seconds compared to $2505$ seconds to get the same parameters. Although the other times are negligible compared to the magnitudes we are dealing with, we have wanted to plot them on this graph (\eqref{fig:benchmark_t_resultsHistGradBoost_GSCV4145_Quantum}). The experiment was done with a $1920$-element database over $4$ columns. With the hyperparameters \textit{param\_learning\_rate} $[0.01 - 1]$, \textit{param\_iter} $[1 - 1000]$, \textit{param\_max\_bins} $[31 - 255]$ and \textit{param\_loss} as categorical $[$squared\_error, absolute\_error$]$ (for more details \eqref{ta1}).

\begin{figure}[]
	\centering
        \includegraphics[width=0.45\textwidth]{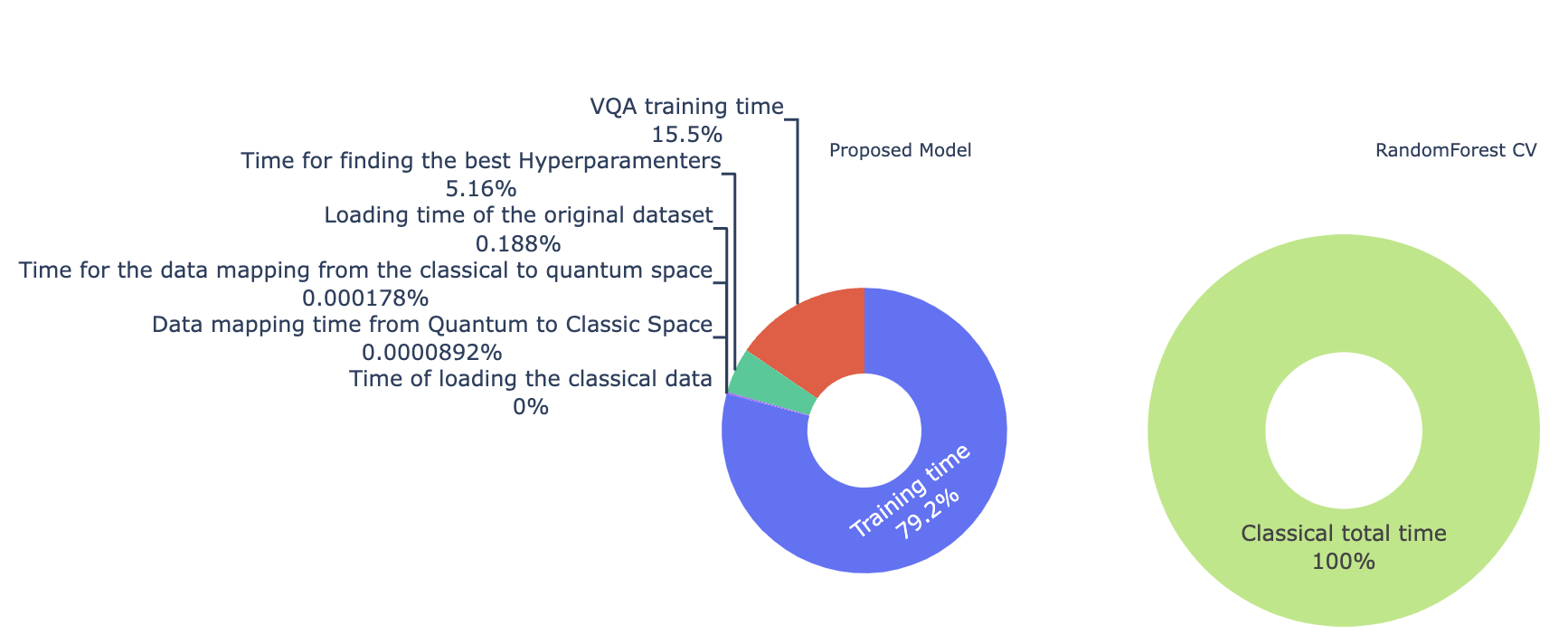}
		\caption{This graph compares the classical model and the hybrid proposed model—the classic model on the right of the image and the proposed model on the left. We are comparing the classically trained data model with the \textit{RandomForest} model using \textit{Grid Search} and \textit{Cross-Validation} with our model. The ratio between the graphs results from saving time in this case, with only a single layer in the quantum circuit. Table \eqref{t_resultsRandomForest_GSCV} analyzes in detail considering the number of layers. In this case, we save $63\%$ of time compared to the classical counterpart. An $88$-element database with $2$ columns considering the following hyperparameters \textit{param\_n\_estimators} $[5 - 250]$ and \textit{param\_max\_depth} $[1 - 15]$ (for more details \eqref{ta1}).}
		\label{fig:benchmark_RandomForestCV3516_Quantum}
\end{figure}

The results of the experiments with the \textit{RandomForest} model are shown by the figure \eqref{fig:benchmark_RandomForestCV3516_Quantum}. The experiment compares the classically trained data model with the \textit{RandomForest} model using \textit{Cross-Validation}. The time saving is $63\%$. For that, the VQA took about $121$ seconds, the search algorithm for the optimal hyperparameters took about $68$ seconds, and the model retraining took about $600$ seconds compared to $2116$ seconds to get the same parameters. The experiment was done with an $88$-element database with over $2$ columns. With the hyperparameters \textit{param\_n\_estimators} $[5 - 250]$ and \textit{param\_max\_depth} $[1 - 15]$ (for more details \eqref{ta1}).

\begin{figure}[]
	\centering
        \includegraphics[width=0.4\textwidth]{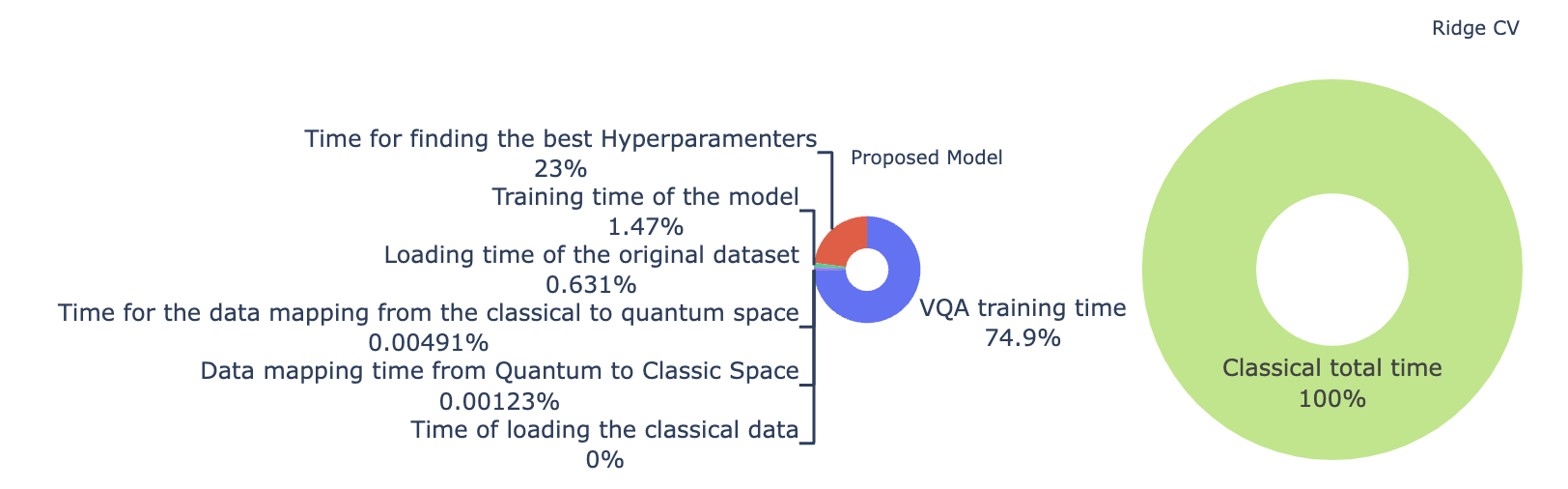}
		\caption{This graph compares the classical model and the hybrid proposed model—the classic model on the right of the image and the proposed model on the left. We are comparing the classically trained data model with the \textit{Ridge} model using \textit{Grid Search} and \textit{Cross-Validation} with the model. The ratio between the graphs results from saving time in this case, with only a single layer in the quantum circuit. Table \eqref{Ridge_GCV} analyzes in detail considering the number of layers. In this case, we can see that we save $73\%$ of time compared to the classical counterpart. A $560$-element database with $3$ columns considering the following hyperparameters \textit{param\_alpha} $[0.0001 - 1]$, \textit{param\_max\_iter} $[1000 - 500000]$ and \textit{param\_solver} as categorical $[\text{svd, cholesky, lsqr, sparse\_cq, sag }]$ (for more details \eqref{ta1}).}
        \label{fig:benchmark_Ridge3435_Quantum}
\end{figure}

\begin{figure}[]
	\centering
        \includegraphics[width=0.45\textwidth]{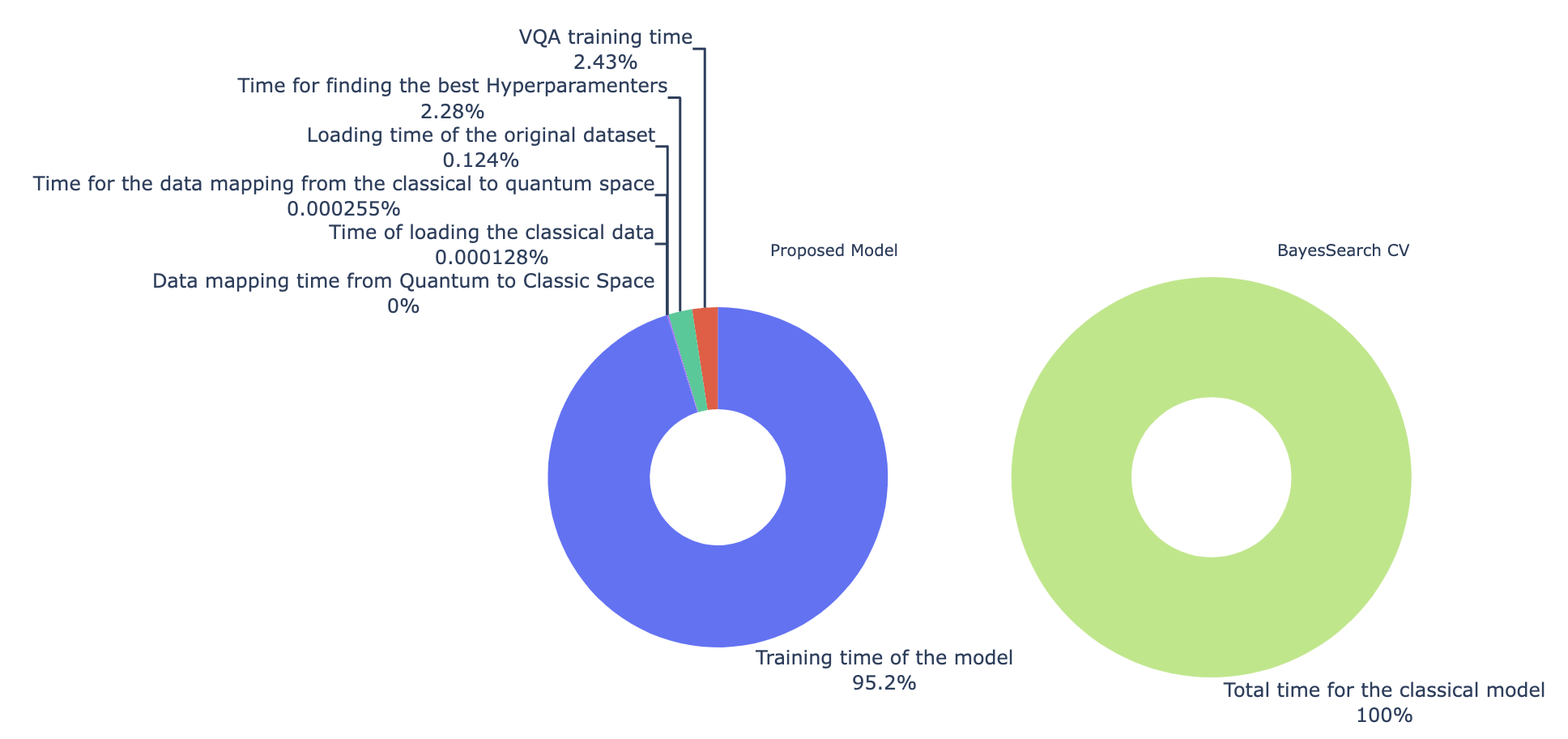}
		\caption{This graph presents the performance of the proposed algorithm compared to its classical counterpart using the \textit{BayesSearch} with \textit{Cross-Validation}. In this case, the proposed algorithm represented 92.5\% of the classical model as a reference which means a 7.5\% time improvement. A $30$-element database with $2$ columns were used, considering the following hyperparameters  \textit{param\_n\_estimators} $[5 - 250]$ and \textit{param\_max\_depth} $[1 - 15]$.}
		\label{fig:benchmark_BayesSearchCV_Quantum}
\end{figure}

The experimentation with \textit{Ridge} model is presented in Figure \eqref{fig:benchmark_Ridge3435_Quantum}, which compares the classically trained data model with the \textit{Ridge} model using \textit{Grid Search} and \textit{Cross-Validation} with our model. The ratio between the graphs represents the savings that we achieve. Here we are saving $73\%$ of the experimentation time.
In this case, the VQA took about $178$ seconds, the search algorithm for the optimal hyperparameters took about $53$ seconds, and the model retraining took about $5$ seconds compared to $2075$ seconds to get the same parameters. The experiment was done with a $560$-element database with over $3$ columns. With the hyperparameters \textit{param\_alpha} $[0.0001 - 1]$, \textit{param\_max\_iter} $[1000 - 500000]$ and \textit{param\_solver} as categorical $[\text{svd, cholesky, lsqr, sparse\_cq, sag }]$ (for more details \eqref{ta1}).
We have realized that for databases classically generated for less than one minute, it is not worth using our algorithm since the setup time in the case of a single hyperparameter is 50 seconds, and for two hyperparameters is one minute and 20 seconds.
Moreover, to verify the proper functioning of our algorithm and how it shares part of its philosophy with \textit{BayesSearch}, we wanted to compare it with \textit{BayesSearch} with cross-validation. As quantum computing is today, the result is quite comparable. The comparison has been simulated on a classical computer. That is, as an example, having the limitations of RAM without the ability to take advantage of quantum parallelization. Our algorithm has been equated by $89\%$ time to the classical\textit{BayesSearch}. Still, we have also experienced cases where \textit{BayesSearch} has exceeded our algorithm by about 100 seconds as the highest observed value. In all cases, our algorithm has found the values of the same hyperparameters as \textit{BayesSearch}. We only compared the algorithm with a single layer. It can be seen in figure \eqref{fig:benchmark_BayesSearchCV_Quantum} the operation at the time and process level of the proposed algorithm. Note how the quantum part goes very fast, and we consume the most time in the retraining of \textit{Bayes Search}. Since the \textit{Bayes Search} with \textit{Cross-Validation} goes back to performing some steps already contemplated in our proposal.

\section{Conclusion} \label{sec:conclusions}
\textit{Hyperparameter tuning} is a research area with great interest in this big-data era. In this article, we have studied using classical and quantum hybrid algorithms to offer a generic solution. We have investigated several scenarios and experiments to propose an efficient model for solving the hyperparameter optimization problem. We have designed an algorithm that fits the current status of quantum computing. Our algorithm and processes can be used in all \textit{quantum-inspired} machines solving real cases in society, waiting for a quantum computer or a system that allows us to have quality service to run it on a commercial quantum computer. Our algorithm has proven robust in all tested scenarios and has given outstanding results. For this reason, we firmly believe that, beyond the hardware limitations and beyond achieving an efficient \textit{qRAM}, if there is a lot of classical data and few functional qubits, algorithms such as the one proposed in this work are the most suitable for the real solutions today.


\section*{Acknowledgements} The authors want to thank Guillermo Alonso de Linaje for the discussions and consideration during the experiments. Also, the authors wish to thank the \textit{Vueling Airlines SA} for allowing the use of their proprietary dataset to perform this work.

\textbf{Compliance with Ethics Guidelines}\\
Funding: This research received no external funding. 

Institutional review: This article does not contain any studies with human or animal subjects.

Informed consent: Informed consent was obtained from all individual participants included in the study.

Data availability: Data sharing is not applicable. No new data were created or analyzed in this study. Data sharing does not apply to this article.



\bibliography{main} 
\end{document}